\documentclass{ieeeaccess}
\usepackage{amsmath,amssymb,amsfonts}
\usepackage{algorithmic}
\usepackage{graphicx}
\usepackage{textcomp}
\usepackage[table,xcdraw]{xcolor}
\usepackage[numbers]{natbib}
\def\BibTeX{{\rm B\kern-.05em{\sc i\kern-.025em b}\kern-.08em
    T\kern-.1667em\lower.7ex\hbox{E}\kern-.125emX}}
\begin{document}
\history{}
\doi{}

\title{Towards trustworthy Energy Disaggregation: A review of challenges, methods and perspectives for Non-Intrusive Load Monitoring}
\author{\uppercase{Maria Kaselimi}\authorrefmark{1}, 
\uppercase{Eftychios Protopapadakis}\authorrefmark{1}, 
\uppercase{Athanasios Voulodimos}\authorrefmark{2} \IEEEmembership{Member, IEEE},
\uppercase{Nikolaos Doulamis}\authorrefmark{1} \IEEEmembership{Member, IEEE}, and \uppercase{Anastasios Doulamis}\authorrefmark{1} \IEEEmembership{Member, IEEE}}

\address[1]{School of Rural and Surveying Engineering, National Technical University of Athens, Athens 15773, Greece.}
\address[2]{School of Electrical and Computer Engineering, National Technical University of Athens, Athens 15773, Greece.}

\tfootnote{This paper is supported by the H2020 European Union Project GECKO ``building GrEener and more sustainable soCieties by filling the Knowledge gap in social science and engineering to enable responsible artificial intelligence co-creatiOn'' funded under the H2020-MSCA-ITN-2020 under grand agreement No. 955422.  }

\corresp{Corresponding author:Maria Kaselimi (e-mail: mkaselimi@mail.ntua.gr).}

\begin{abstract}
Non-intrusive load monitoring (NILM) is the task of disaggregating the total power consumption into its individual sub-components. Over the years, signal processing and machine learning algorithms have been combined to achieve this. A lot of publications and extensive research works are performed on energy disaggregation or NILM for the state-of-the-art methods to reach on the desirable performance. The initial interest of the scientific community to formulate and describe mathematically the NILM problem using machine learning tools has now shifted into a more practical NILM. Nowadays, we are in the mature NILM period where there is an attempt for NILM to be applied in real-life application scenarios. Thus, complexity of the algorithms, transferability, reliability, practicality and in general trustworthiness are the main issues of interest. This review narrows the gap between the early immature NILM era and the mature one. In particular, the paper provides a comprehensive literature review of the NILM methods for residential appliances only. The paper analyzes, summarizes and presents the outcomes of a large number of recently published scholarly articles. Also, the paper discusses the highlights of these methods and introduces the research dilemmas that should be taken into consideration by researchers to apply NILM methods. Finally, we show the need for transferring the traditional disaggregation models into a practical and trustworthy framework. 
\end{abstract}


\titlepgskip=-15pt

\maketitle

\section{Introduction}
\label{sec:introduction}
\PARstart{E}{nvironmental} policies, responses or solutions to climate change at global scale is prerequisite to raise awareness of individuals or social groups on protecting our world and retaining its sustainability \cite{Rolnick2019}. There are various ways that householders could contribute towards a sustainable living. One of them is by {\it reducing their energy consumption}. To this end, such a reduction requires a change of the energy-related humans' behavior in their household. To shape this behavioral change, {\it consumers need to become aware of the energy they consume}. However, end-consumers often lack knowledge about potential energy savings, existing policy measures and relevant technologies. Most household consumers are only aware of general information related to their consumption through monthly electricity bills. Nonetheless, the effectiveness of feedback on energy consumption is crucial and is usually translated into good practices and tailored advice for energy saving.

\begin{figure*}
\centering
\includegraphics[width=\textwidth]{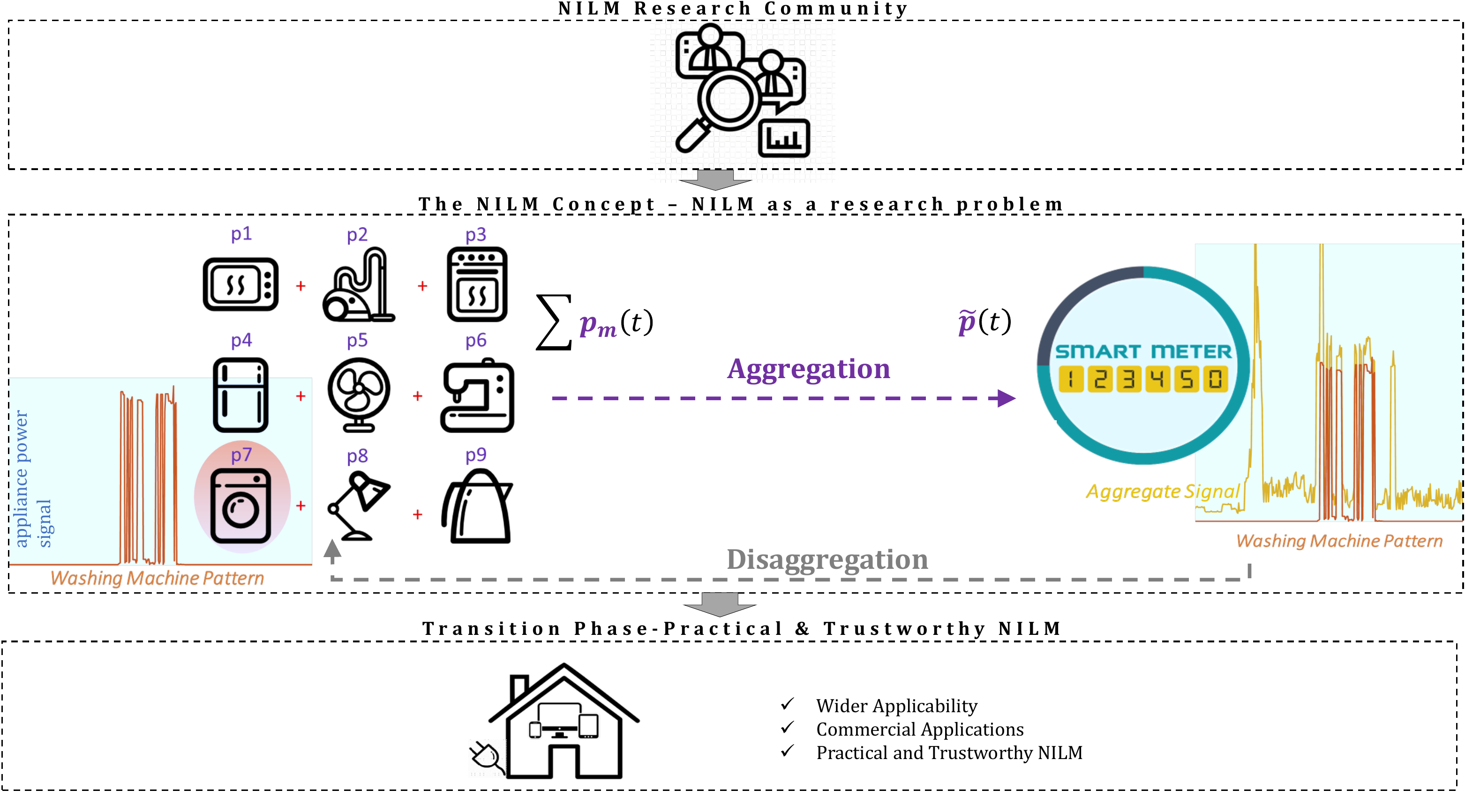}
\caption{\textbf{The energy disaggregation concept.Until recently, the most crucial issue was to create NILM algorithms with good performance. NILM belongs on the category of inverse problems and formulating this mathematical problem and adapt it using machine learning models was quite challenging (early immature NILM era). Nowadays, given that the state-of-the-art algorithms achieve good performance, we run across the transition phase where the research interest is concentrated on practical and trustworthy NILM algorithms. }}
\label{fig:nilmproblem}
\end{figure*}

Non-Intrusive Load Monitoring (NILM) uses the aggregate power signal of a household as input to estimate the extent to which each appliance contributes to the aggregate energy consumption signal \cite{Hart1992}. Using NILM techniques, one can provide itemized energy bills and personalized energy savings recommendations. Thus, NILM is an efficient and cost effective framework for \textit{energy consumption awareness}. Power disaggregation is applied to enhance awareness on the energy consumption behavior of consumers in the household and therefore guide them towards a prudent and rational utilization of energy resources \cite{Najafi2018}. 

NILM techniques combined with Home Energy Management Systems (HEMS) and Ambient Assisted Living (AAL) technologies, assist on decisions for efficient energy management \cite{ruano2019nilm}. For the successful cooperation of these technologies \cite{bousdiat2020augmenting}, \cite{hernandez2019applications}, scalable NILM methods are necessary, since  they address issues arising from different user behaviors with respect to appliance usage \cite{murray2018appliance}. However, there are concerns about privacy and security of the consumers' sensitive data (see Section \ref{sec:challenges}). In literature, there is an early attempt to deliver privacy-preserving and personalized NILM using federated learning technologies \cite{Zhang2021fednilm}. However, additional research work in the field should be conducted. Additionally, NILM is also used for demand side response scenarios. Demand response provides the possibility of shifting demand away from the peak and thus decreasing the
corresponding cost of energy \cite{Kelly2015}, \cite{salani2020non}.
In this context, energy disaggregation permits power supplier companies to identify a device with a high consumption rate at a peak hour in a household and sends a message to the corresponding users asking them to postpone their usage to smooth out the current peak in the demand. In this context, one can consider the works of \cite{berges2008training} and  \cite{lucas2019load}. 

Another usage of the disaggregated electrical consumption is to identify malfunctioning appliances. As an instance, NILM approach is applied to detect the frosting cycle of a fridge with a damaged seal, which is more frequent than the normal one \cite{Najafi2018}. For instance, the possibility of utilizing NILM for anomalous behavior detection has been addressed in \cite{rashid2019can}, where the adoption of denoising autoencoders is considered in \cite{Bonfigli2017}. The application of NILM in home appliance malfunctions detection is essential to detect problems in appliances' operation which can be very useful for improving the concept of self-monitoring appliances since the latter can identify potential problems in their operation and send related messages or take actions \cite{Kelly2015}.  

Finally, NILM solutions can be properly combined with eXplainable AI (XAI) techniques to be beneficial for the user to realise demand-response scenarios. XAI provides interpretable AI models which are capable of explaining their outcomes and provide trust on their performance \cite{samek2019explainable}. Thus, we need to point out trustworthy aspects of NILM solutions and increase the acceptance of these tools by the society, revealing potential perspectives and limitations. Here, explainable AI techniques combined with visual analytics could be beneficial for the user to realise demand-response scenarios.  

The scope of this work is summarized as follows:

\begin{itemize}

    \item{provides a short literature review on the existing NILM methods for residential appliances, and highlights the trustworthiness aspects of the current state-of-the-art methods.}
    
    \item{collects the research dilemmas that are appeared in literature for solving the NILM problem, and discusses the advantages and disadvantages between the different approaches.}   
    
    \item{highlights the existing challenges in NILM and discusses the barriers and limitations towards a reliable, practical and trustworthy NILM framework.}
    
    \item{discusses the future perspectives on NILM models under a trustworthy framework.}
    
\end{itemize}

The remaining survey is structured as follows: In Section \ref{sec:back}, we define the topic of this literature review and describe the NILM problem, identifying the relevant gaps and challenges in the current knowledge. Section \ref{sec:review} synthesises the information in literature about NILM into a summary, organised from a chronological point of view. The brief literature review identifies the important works in energy disaggregation area, starting from the early NILM era, until nowadays, when NILM researchers approached advanced NILM issues and challenges using state-of-the-art signal processing and machine learning algorithms. The remaining sections are organized to follow the general pipeline of NILM in the literature, that is: a) signal pre-processing techniques and feature extraction and selection, b) the machine learning part of the algorithm, c) load disaggregation and evaluation of the results. In particular, Section \ref{sec:signal} is an in-depth analysis of the common signal pre-processing and feature extraction techniques available in literature for NILM. Section \ref{sec:dl4nilm} identifies the opposing views in machine learning techniques applied for NILM and identifies the trends emerging from the analysis and authors understanding in NILM. Section \ref{sec:trust} constitutes an important dimension in this literature review. This section is a discussion about the trustworthiness of the NILM algorithms. Trustworthy AI has attracted immense attention recently, allowing humans to realize the full potential of AI, so that humans can fully trust and live in harmony with AI technologies. In this literature review, we discuss the key papers towards trusted NILM solutions and we identify the future perspectives in NILM in order to build upon trustworthiness. Section \ref{sec:evaluation} is a summary of the existing datasets, evaluation metrics and open NILM tools. Finally, Section \ref{sec:discussion} is a discussion and a final conclusion about NILM techniques, towards efficient and trustworthy NILM algorithms during the whole NILM implementation pipeline. 

\section{Background on NILM}
\label{sec:back}
\subsection{NILM problem formulation}
Disaggregation of households’ power consumption allows grid operators to improve their predictions in energy demand and is an important part of providing a stable supply of power to all customers on a power grid \cite{Dinesh2019}. The
consumption profile of appliances is identified through disaggregation and then, the obtained appliance-level
load profiles along with meteorological information are employed to predict the future usage, as in \cite{Basu2013} and \cite{Dinesh2019}.

Thus, energy disaggregation is of great importance for energy conservation and planning. Given the power consumption per appliance the forward problem is to predict the total power consumption in a household (see Fig. \ref{fig:nilmproblem}). Energy disaggregation is described as an inverse ill-posed problem which aims to estimate unknown individualized components from aggregate measurements. We assume the aggregate signal ${p}(t)$ at a discrete time index $t$ to be equal to the summation of the individual appliances' power consumption ${p_m}(t)$ plus an additive noise $\epsilon(t)$. Thus: 
\begin{equation}
{{p}(t)=\sum_{m=1}^{M} {p_m}(t)+\epsilon(t)}
\label{eq:1}
\end{equation}

In Eq. (\ref{eq:1}) variable $m$ refers to the $m$-th out of $M$ available appliances. Under a NILM framework, the individual appliance power consumption ${p_m} (t)$ is not a priori available, assuming the absence of installed smart plugs. Instead, only ${p}(t)$ is given. The inverse ill-posed problem, called NILM (see Fig. \ref{fig:nilmproblem}), is to calculate the best estimates ${\hat{p}_m} (t)$ of the actual values of the appliance power consumption $p_m (t)$, given the aggregate power value  ${p}(t)$.

\subsection{Challenges on NILM}
\label{sec:challenges}
Various approaches have been proposed to solve the NILM problem, as presented in Section \ref{sec:review}. Some of the most successful ones exploit deep learning neural network structures for modelling an energy disaggregation problem (e.g., \cite{Kaselimi2019a}). Nevertheless, non-intrusive load monitoring is a challenging task. We hereby provide an indicative list of the NILM challenges based on our current understanding of the field. Some of them are well-studied, whereas others are immature and there is an ongoing research in these topics. 

\noindent
\textit{ ``Challenge 1: To create reliable algorithms with good generalization ability'' }

Most state-of-the-art techniques have not been applied successfully in unseen houses (transferability), across different households and datasets \cite{Murray2019}. Therefore, it is difficult to create reliable algorithms with a good generalization ability. Large scale trials is a necessary step towards this direction. Another aspect of NILM problem is to create robust models and deal with noisy datasets and appliances with abnormal behavior. Noisy aggregate energy consumption measurements significantly deteriorate the performance of NILM methods. In addition, a common problem in NILM is that the targeted appliances have unsteady signatures or present abnormal behavior. On top of these barriers, inadequate datasets deteriorate overall models' performance \cite{DIncecco2019}.

\noindent
\textit{ ``Challenge 2: To develop hybrid NILM models incorporating user's feedback and techniques that support continuous learning'' }

Consumers' habits and seasonality significantly affects the energy usage patterns and introduces an additional challenge in load monitoring. Various factors, including environmental, socioeconomic factors, etc., affect the operation of various domestic appliances. Users' feedback in NILM algorithms is crucial in order to improve the models' accuracy. Modern NILM methods should be dynamically updated and improved based on user’s recommendations.

\noindent
\textit{ ``Challenge 3: To provide explainable NILM models with reasoning behind the model estimations'' }

Even though the recently proposed models in literature provide competitive accuracy, the inner workings of these models is less clear. Understanding and trusting the outputs of the networks help in improving the designs, highlights the relevant features and aspects of the data used for making the decision, provides a better insight of the model accuracy, and also inherently provides a level of trust in the value of the provided consumption feedback to the NILM end-user. 

\textit{ ``Challenge 4: To achieve fairness in NILM'' }

The various socioeconomic, environmental, etc. factors that affect power load consumption, lead to multiple distinct data distributions (e.g., geographic groups, or social categories) that should be expressively modelled and represented under a NILM framework. Thus, the NILM AI framework learns to predict outcomes that are accurate with respect to the ground truth data of the target appliance used for validation, but also fair with respect to a set of pre-defined fairness metrics, leveraging sufficient and diverse training data. Except for fair data and models, fair performance evaluation, that enables proper benchmarks is another important aspect for practical NILM.

\textit{``Challenge 5: To provide privacy-preserving outcomes using secure NILM models'' }

To achieve the real-world applicability of NILM, we should previously address privacy concerns in NILM applications in order to provide personalized NILM services. The emerging NILM deep learning models require for massive amounts of real-life data to improve their performance. Thus, data security and user privacy has become an important issue.

\section{A Brief NILM Literature Review}
\label{sec:review}
Hart \cite{Hart1992} first introduced the NILM as a method capable of estimating the energy used by individual appliances, given only the total energy consumption. At first, NILM is modeled as a linear combination problem, where each time the algorithm estimates the percentage of the  total power consumption that an active appliance consumes. The ability of collecting massive amounts of data related to household power consumption, among with the evolution of deep learning methods, made possible the NILM formulation as a non-linear problem. Thus, we have observed the pairs of the data ($p_m(t), p(t)$) where $p_m(t)$ and $p(t)$ denote respectively the power reading of an appliance and the mains at time $t$. Given that there are plenty of observations, it is possible to train learning models to represent the relationship between $p_m$ and $p$ \cite{DIncecco2019}. Since then, a number of studies have extended the previously simple linear model into a non-linear one, applying various deep learning schemes.

\subsection{The early NILM era [1995--2014]} 
\label{subsec:early_Nilm}
Hart was the first to propose a method for disaggregating electrical loads through clustering of similar events based on appliances’ characteristics \cite{Hart1992}. This approach employed Combinatorial
Optimization (CO), which at the time was the standard technique for disaggregation problems. This first approach had a major shortcoming: Combinatorial
Optimization performed the power disaggregation on each instant independently of the others, without considering the load evolution through time. Most common approaches to solve the NILM problem are based on unsupervised event detection in the aggregate signal, whereas supervised classifiers are used to assign known appliances to detected events in order to estimate the power trace of individual appliances. 
Different classification tools have been widely used, including Support Vector Machines (SVM) \cite{altrabalsi2016low}, neural networks, Decision Trees (DT) \cite{liao2014non}, and hybrid classification methods \cite{He2016}, \cite{lin2014non}. Contrary to the aforementioned classic methods, other methods such as Dynamic Time Warping (DTW) are used for comparing and grouping windows from daily profiles and identifying unique load signatures \cite{elafoudi2014power}. The main object of controversy in these approaches refers to the difficulty of classifying multi-state appliances \cite{lin2014non}, \cite{Kim2017}. Multi-state appliances require a long range pattern to be trained for their detection \cite{Kim2017}. Graph Signal Processing (GSP) \cite{He2016} is a concept that effectively captures spatio-temporal correlation among data samples by embedding the structure of signals onto a graph. Zhao et al. \cite{zhao2018improving} propose a low-resolution, event-based, unsupervised GSP approach. Recently, a Modified Cross-Entropy method for event classification has been suggested \cite{machlev2018modified}, which is based on CO and formulates NILM as a Cross-Entropy problem. 

Hidden Markov Models (HMM) and various extensions of them are advocated in order to explore the possible combinations among the different appliances’ state sequences \cite{kolter2012approximate}, \cite{Bonfigli2017}, \cite{Kong2018}, \cite{bajovic2018optimal}. In this light, HMMs are state-based, so the studied appliances should have discrete states in their signatures \cite{mauch2015new}. As the number of appliances increases, the number of combinations of states sequences is increased exponentially, increasing respectively problem’s complexity \cite{mauch2015new}. In addition to this, time complexity is also increased, leading to the reduction of model’s classification performance \cite{Kim2017}. Makonin et al. \cite{makonin2015exploiting} have proposed a super-state HMM and a sparse Viterbi algorithm in order to reduce the complexity. Another limitation  of HMM-based approaches is that they tend to fail in the presence of unknown appliances \cite{mauch2015new}. Rahimpour et al. \cite{Rahimpour2017} proposed a matrix factorization technique for linear decomposition of the aggregated signal using as bases of this learned model the appliances’ signatures resulting in an efficient estimation of the energy consumption per appliance.

\subsection{Deep learning based NILM [2015--2019]}
\label{subsec:deep_learning}

NILM algorithms received renewed attention, mostly thanks to the increased number
of datasets stemming from smart electric meters installed in domestic residences (\cite{Makonin2013}, \cite{Murray2017}), and thanks to the increased number
of these datasets, the proposed solutions to NILM shifted to a supervised learning process. With the rise of deep learning, a new family of methods have been introduced that exploit deep neural network structures to solve the ill-posed NILM problem. Deep learning techniques have been applied mostly to low frequency NILM approaches since 2015 \cite{Kelly2015}. 

A common approach is to treat the aggregated signal as a corrupted by noise signal of an appliance. Under this view, denoising autoencoders (DAE) are excellent techniques used to reconstruct a signal from its noisy version. This architecture has been initially proposed by Kelly and Knottenbelt \cite{Kelly2015}, while others expand the idea proposing alternative DAE architectures, such as \cite{Bonfigli2017}.

Exploiting the temporal character and dependencies of the power signal, another family of deep learning models, Recurrent Neural Networks (RNN), have proved efficient under the NILM framework. Here, NILM is treated as a supervised learning problem with times series. RNNs and their variants, such as Long Short-Term Memory Networks (LSTM) and Gated Recurrent Units (GRU) have been primarily used, as they are very popular and effective with 1D time series data.  Relevant studies have been carried out in the past (\cite{Murray2019}, \cite{Kelly2015}, \cite{Kim2017}). In a previous work of ours, we have also proposed a Bayesian optimized bidirectional LSTM model for NILM \cite{Kaselimi2019a}, whereas in \cite{Kaselimi2020} a context-aware LSTM model adaptable to external environmental conditions is presented.

Although Convolutional Neural Networks (CNN) are traditionally developed for two-dimensional imagery data (\cite{voulodimos2018deep}), one-dimensional CNN can be used to model the temporal character of sequential timeseries data. Few researches \cite{Kaselimi2019b} have tried to enrich CNN structures providing a recurrent character, such as CNN-LSTM and Recurrent Convolutional Networks. In \cite{Harrell2019} a causal \textit{1}D convolutional neural network for NILM is proposed. Others introduce the concept of data sequences \cite{Kelly2015} to feed the classic structure with historical past values of power load. Others \cite{Zhang2016} propose a sequence to point CNN architecture, underscoring the importance of sliding windows to handle long-term timeseries. Alternatively, sequence to sequence architectures have also been proposed \cite{Chen2018}.

\subsection{Current Advancements in NILM [2020-- ]}
\label{subsec:advancments}
Recently, there are various advanced machine learning methods applied for NILM. These methods do not only provide competitive accuracy against the traditional NILM methods, but also propose possible solutions to solve the remaining challenges in NILM and is an attempt towards a trustworthy NILM in terms of accuracy, robustness, reliability, explainability and fairness. Some of the worth mentioned works are presented here. 

Generative adversarial networks (GANs) recently have been applied for NILM. An early attempt for solving NILM using a GAN-based framework is adopted in \cite{Bao2018}. Then, Kaselimi et al. \cite{Kaselimi2020b} propose a generative adversarial network for sequence to sequence learning, whereas Pan et al. \cite{Pan2020} achieve sequence to sub-sequence learning with conditional GANs. Chen et al. \cite{Chen2019} propose a context aware convolutional network for NILM that has been trained adversarially. Most of these studies exploit the robustness to noise that the adversarial training process achieves. 

Transformer models were explored as alternative architecture for neural machine translation tasks within the past two years \cite{wang2020minilm}. Recently, a transformer based architecture that utilizes self-attention for energy disaggregation has been adopted by \cite{yue2020bert4nilm}, to handle power signal sequential data. 

Given that, most of the existing deep learning models for NILM use a single-task learning approach in which a neural network is trained exclusively for each appliance. In contrast to a single-task learning approach, the work of \cite{faustine2020unet} proposes UNet-NILM for multi-task appliances' state detection and power estimation, applying a multi-label learning strategy and multi-target quantile regression. The UNet-NILM is a one-dimensional CNN based on the U-Net architecture initially proposed for image segmentation.

Explainable AI (XAI) attempts to promote a more transparent and trust AI through the creation of methods that make the function and predictions of machine learning systems comprehensible to humans, without sacrificing performance levels \cite{vilone2020explainable}. Explainable NILM networks proposed by \cite{murray2020explainable}, try to understand the inner workings of the machine learning models used for NILM.

\section{Signal analysis and feature extraction}
\label{sec:signal}
It is experimentally proven that applying data re-sampling, data cleaning methods and dataset balancing, significantly improves energy disaggregation in terms of accuracy and generalisation abilities \cite{klemenjak2020towards}.However, NILM techniques are relatively immature at this stage, and have not reached the point where best practices can be defined. Thus, this section summarizes the most common practises and methods available in literature and discusses the advantages and disadvantages of these methods based on authors understanding in the field. 

\subsection{Outline of the existing practices for NILM data pre-processing}

\subsubsection{Balancing}
A large amount of paradigms of an appliance in operation is necessary for the supervised learning algorithms to be able to detect the appliance in the total power signal with good accuracy. However, it is observed that for some of the appliances, the switch-on times (active time) are relatively small compared to switch-off times (idle time). For example, an espresso machine is open only for a few minutes every day, thus it is difficult to collect a large amount of representative paradigms where the appliance is on. In addition, the monitoring of load consumption in households reveals significant differences of individual habits and daily routines of occupants. These habits and individual routines affect the usage of household appliances and therefore, the number of events found in energy consumption data. 
The existing datasets in NILM, are characterized as highly imbalanced. However, data balancing improves the models performance and alleviates overfitting at the same time. There have been observed two different kinds of imbalance in NILM datasets: (i) the imbalance that is caused by the difference in the active and idle time of appliances and (ii) the imbalanced appeared because some appliance types are represented
by more measurements than others (e.g., espresso machine versus air condition). Here, we emphasize that the majority of the commonly used data sets have limited time duration, thus the available training samples are few, and usually, this has impact in the model's performance (see Section V-II).  

There are various research works dealing with the first imbalance case caused by the difference
in the active and idle time of appliances \cite{de2017handling}. This can influence the performance achieved by a particular
classifier trained using these data \cite{de2017handling}. Balanced data are necessary in order to avoid the issue of bias due to lack of adequate number of appliance activations, which is a common problem in many NILM datasets. Every appliance should have a representative number of examples of its activation in the training learning process for supervised disaggregation algorithms. For the second imbalanced case, where some appliance types are represented
by more measurements than others, this could be a problem in cases of "all-in-one" models. Different techniques for handling this
imbalance and avoiding biasing the classifiers during training are
investigated in \cite{de2017handling}. 


\subsubsection{Handling sample rates and missing data}
\label{subsec:handling_sample_rates}
High-frequency energy meters are essential in order to capture the transient events or the electrical noise generated by the electrical signals \cite{zoha2012non}. Thus, the more frequently energy consumption is measured, the more detailed is the captured information of energy consumption. However, increasing the sampling frequency will increase the data to be stored, processed, or transmitted which in turn increases the hardware cost exponentially  \cite{liao2014non}, \cite{schirmer2020energy}. Therefore, most recent studies focus on low sampling frequency data, as the majority of commercial smart meters collect data usually at 0.1 Hz or up to 1 Hz to minimize the hardware cost of smart meters, their financial cost and to address the transmission and data-storage capacity limitations \cite{schirmer2020energy}.

Most of the data sets come in a variety of sampling rates \cite{klemenjak2020towards}, thus, in order to propose a robust NILM model that incorporates information from different data sets, it is important to have a tool to successfully re-sample the data. Using re-sampling techniques, problems related with missing data were overcome and with a down-sampling the overall size of a data set was reduced, targeting to a more flexible data inputs  \cite{pereira2019dscleaner}. Data re-sampling filters out erroneous readings and finds the gaps in data readings which are necessary practices in order to improve the models performance.


\begin{figure}
    \centering
    \includegraphics[width=8.6cm]{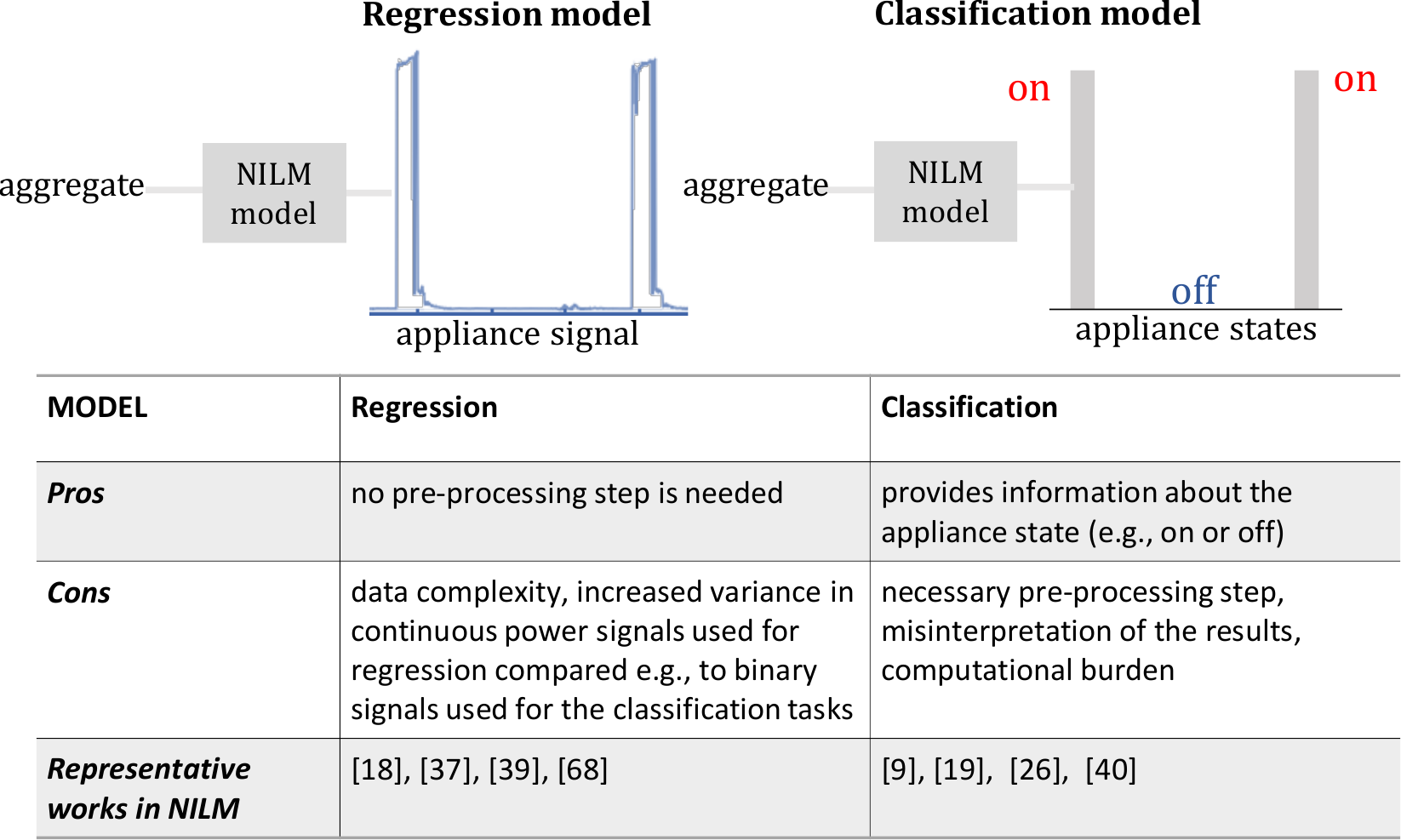}
    \caption{\textbf{NILM as a regression or classification problem.} }
    \label{fig:regcl}
\end{figure}

\subsubsection{Optimal features extraction and selection}
\label{subsec:opt_feat_extr}

Feature selection is an essential step in machine learning in which a subset of relevant features or variables is identified and selected to be used in the model construction. Usually, the various values and features are selected according to their statistical importance determined by various algorithms (such as the ReliefF algorithm \cite{bao2020feature}), resulting to a meaningful feature vector. There are various studies that experimentally approve that information which is offered by more (additional) features could improve the accuracy and reliability of the NILM algorithms \cite{Kaselimi2019b}, \cite{Harrell2019}. However in most of the open access datasets, this information is not always available. The different features that can be extracted from the acquired data are determined based on the sampling rate of power meters that is {\it low-frequency or high-frequency.} 

\textbf{Low-frequency measurements:} Some of the commonly used low-frequency features for load identification are the active (P) -reactive (Q) power plane (P-Q plane) \cite{Barsim2013AnAF}, macroscopic transients, active power \cite{dinesh2016non} and current and voltage-based features \cite{hassan2013empirical}, \cite{Harrell2019}. Here, we highlight that although there are few works dealing with additional features for low-frequency machine learning techniques, the majority of the approaches focus solely on active power measurements, which is a variable that exists in the majority of the open access data sets. Also, there are a few studies where a set of features based on active power values is extracted. These features could include -but not limited to- various statistical measures, such as minimum, maximum, mean and median values, percentiles, standard deviation, skewness, kurtosis, etc. 

\textbf{High-frequency measurements:} As regards to the high-frequency steady state and transient features used in load identification \cite{sadeghianpourhamami2017comprehensive}, examples are spectral envelope, wavelets, shape features, raw wave-forms, Voltage-Current (V-I) trajectory, etc. High-frequency based NILM methods found in the bibliography are based either on spectrogram analysis \cite{wu2019concatenate}, \cite{zoha2012non} or on current-voltage trajectories \cite{schirmer2021double}. Rather than relying solely on time‐domain analysis, in \cite{himeur2021intelligent} a two‐dimensional (2D) representation is used as description of the power signal. Also, in \cite{wu2019concatenate}, high-frequency current data is converted to spectrograms by Short Time Fourier Transform (STFT) and set as the model input. Furthermore, in \cite{liu2018non} a V-I trajectory enabled transfer learning method for NILM is proposed. At first, the V-I trajectories are transformed to visual representation in a color space, and then, a pretrained convolutional neural network is fine-tuned to perform classification on the color images of V-I trajectories. A comprehensive review that highlights the dependence between the NILM features and the sampling rate used is provided by \cite{ruano2019nilm}.

\section{Machine learning for NILM}
\label{sec:dl4nilm}
The machine learning approaches in solving NILM problems should always keep the trade-off between the complexity of the model/architecture and the accuracy improvement. Here, we introduce a list of common dilemmas that the NILM researchers face. The solution to these dilemmas is not always obvious and is depended from the data acquisition, datasets availability and accessibility, the need for near real-time NILM capabilities, the system's scalability as well as whether the system is able to recognise various different appliance and types of appliances \cite{Nalmpantis2018Machine}. 

\subsection{Research dilemmas and Conflicting views}

\subsubsection{Classification or Regression model}
\label{subsec:class_or_reg}

In the majority of NILM datasets, both the aggregated power load and the power consumption of each monitored device are included. On the contrary, the appliance switch on events are provided only in a few datasets \cite{kelly2015theukdale}. Thus, a regression problem to predict the consumption of each device is naturally derived from the data \cite{precioso2020nilm}. However, most works in NILM address the classification problem of determining whether the appliance is in operation or not, rather than estimating its consumption at each time interval. The advantages and disadvantages of the two approaches in NILM are summarized in Fig \ref{fig:regcl}. 

A classification problem (usually also naming as event-based approaches in the literature \cite{zoha2012non}) requires a threshold or even more sophisticated event detection procedures to determine the appliance state given the continuous power load. An event detection in NILM aims to detect the times when state transition actions occur in the power consumption signal. The state transition actions normally include appliance turn-on, turn-off, speed adjustments, and function/mode changes. The event detection becomes more challenging when the appliances with the different level of energy demand are operating simultaneously, requiring high sampling data to create unique signatures and to differentiate one appliance from the others. Moreover, keeping track of on/off timestamp, duration of on/off, and calculation of average load consumption during specific active periods makes the algorithms more computationally burdened \cite{ayub2020multi}. In classification techniques, an accurate event detection approach is prerequisite for precise load identification and valid power consumption estimation. Depending on how this pre-processing step is performed, the performance and interpretation of the final results may vary in a significant manner. 

Defining NILM as a regression problem obviates the intermediate step of event detection (non-event based approaches) and the per-appliance disaggregation value is obtained directly from the results of the regression output layer.

\subsubsection{Multi-target or single-target model}
\label{subsection:multitar}

\begin{figure}
    \centering
    \includegraphics[width=8.6cm]{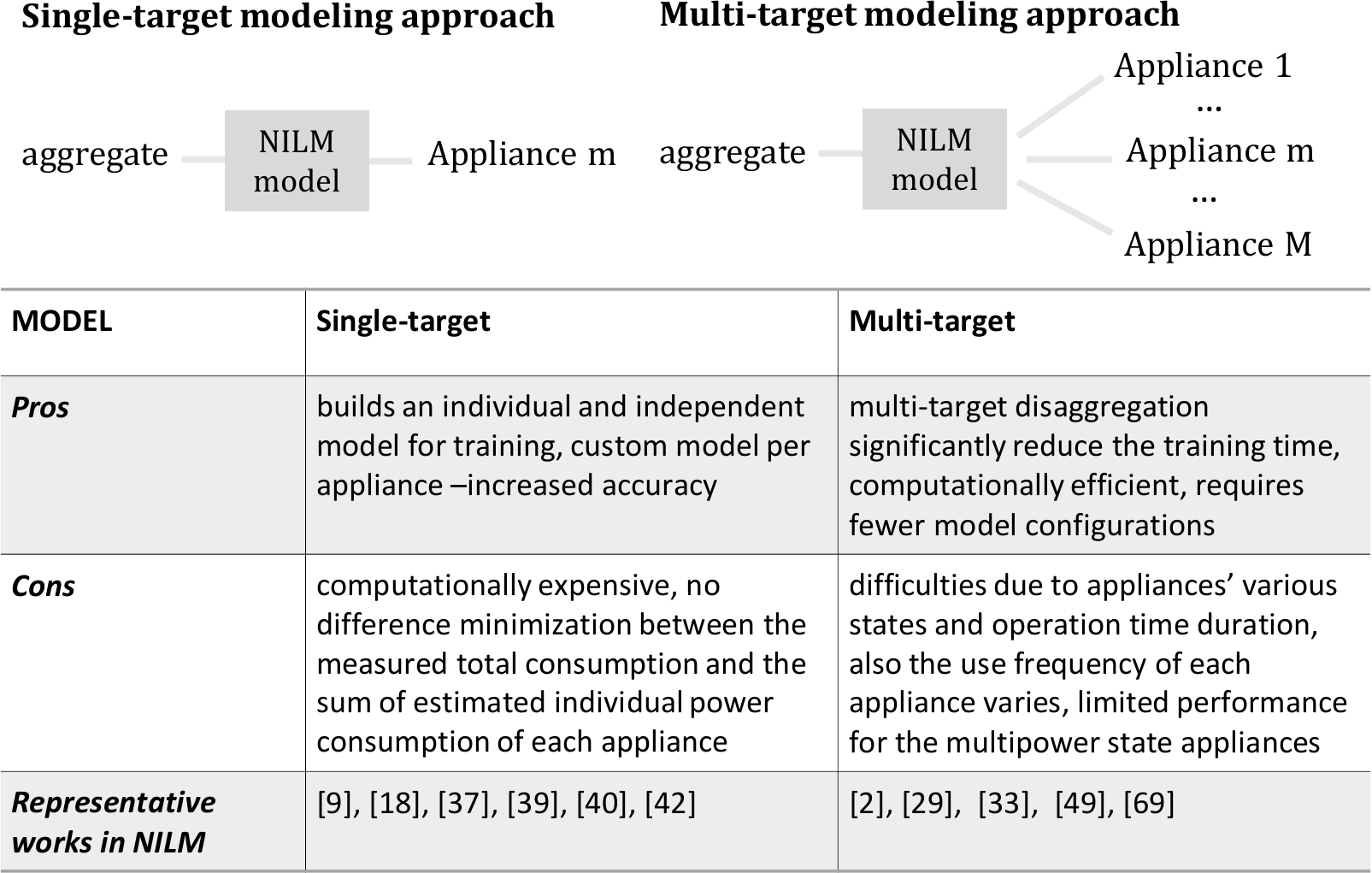}
    \caption{\textbf{The single or multi target modeling approach dilemma in NILM.}} 
    \label{fig:singlemulti}
\end{figure}

A disaggregation model can be trained either as a single target \cite{jiang2021deep}, \cite{Kaselimi2020} or multi-target \cite{xia2020non} regression problem, or as a
single-label \cite{jorde2018electrical}, \cite{devlin2019non} or multi-label \cite{kim2019appliance}, \cite{li2018residential}, \cite{singhal2019simultaneous} classification problem. Single- and multi-target NILM classification methods were explored in many works in the literature, to our best knowledge. Instead, multi-target regression models for disaggregation are not well-studied yet. It is worth mentioning, that in the work of \cite{faustine2020unet}, the authors propose a multi-target model based on U-Net architecture that simultaneously performs multi-task classification and regression.

The early works in NILM aim at multi-target classification models (see \cite{Hart1992}, \cite{kolter2012approximate}, \cite{makonin2015exploiting}). The main difficulty arising from the multi-target approach is that the pattern and behavior of each appliance differs so, it is difficult to create a unique model able to disaggregate the main power consumption of a household simultaneously for all the individual appliances (see Fig. \ref{fig:singlemulti}). This is due to the fact that there are appliances with various states and operation time duration. Also, the use frequency of each appliance varies. Given that these appliances have different frequency of appearance, it is difficult to create the  ``universal'' balanced dataset needed to fit each appliance's needs. This explains why most of these models are usually formulated as sparse models \cite{makonin2015exploiting}, \cite{singh2019non}, \cite{kolter2010energy}, in order to handle the rare appliance activation events in time. Sparsity is a common problem in NILM, because the time duration that most of the appliances are in operation is relatively small compared to the time duration that the appliance is off.

Later, with the release of large amounts of datasets, supervised machine learning models that represent the non-linear relationship between the aggregate signal and a single appliance each time (single-target model), became a trend method for solving NILM. Nowadays, most of the existing deep neural network models for NILM use a single-task learning approach in which a neural network is trained exclusively for each appliance. On the one hand, this is an efficient approach since the analytical models for each appliance can be developed independently of each other and transferred to unseen houses. On the other hand, in order to perform a full disaggregation into a single house, a number of different models should be trained at first and then, be activated in order to detect how many appliances are in operation for a specific time. Also, these techniques need for a vast amount of data for training and houses equipped with smart plugs per appliance. The challenge here is to relax NILM algorithms, and  propose models that require less amount of data leveraging, for example, the advantages of semi-supervised techniques. In addition, these methods can significantly under- or over- estimate the aggregate power consumption since they do not minimise the difference between the measured total consumption and the sum of estimated individual power consumption of each appliance. This happens because in a single-target approach, each model is trained independently from the other appliances \cite{he2019ageneric}. Recently, the work of \cite{faustine2020unet}, suggests a multi-target U-Net network with promising performance against the traditional single-task learning, whereas in \cite{Buddhahai2021anon} a multi-target NILM algorithm is proposed using a random forest regressor.

\subsubsection{Supervised or unsupervised learning}

NILM systems are categorized into supervised and non-supervised approaches depending on whether or not they require a training process prior to the model deployment on a target household. Unsupervised NILM systems do not require training and, therefore, it is expected to have a wider applicability. The early works in NILM mostly targeted to unsupervised learning \cite{Barsim2013AnAF}, as the labeled datasets were limited. These works where mostly use an event detector -clustering algorithm, and then, a transition matching stage follows, in which the on- and off-events belonging to the same appliance are grouped together so that the whole operation interval of each appliance can be inferred \cite{Barsim2013AnAF}. 

\begin{figure}
    \centering
    \includegraphics[width=8.6cm]{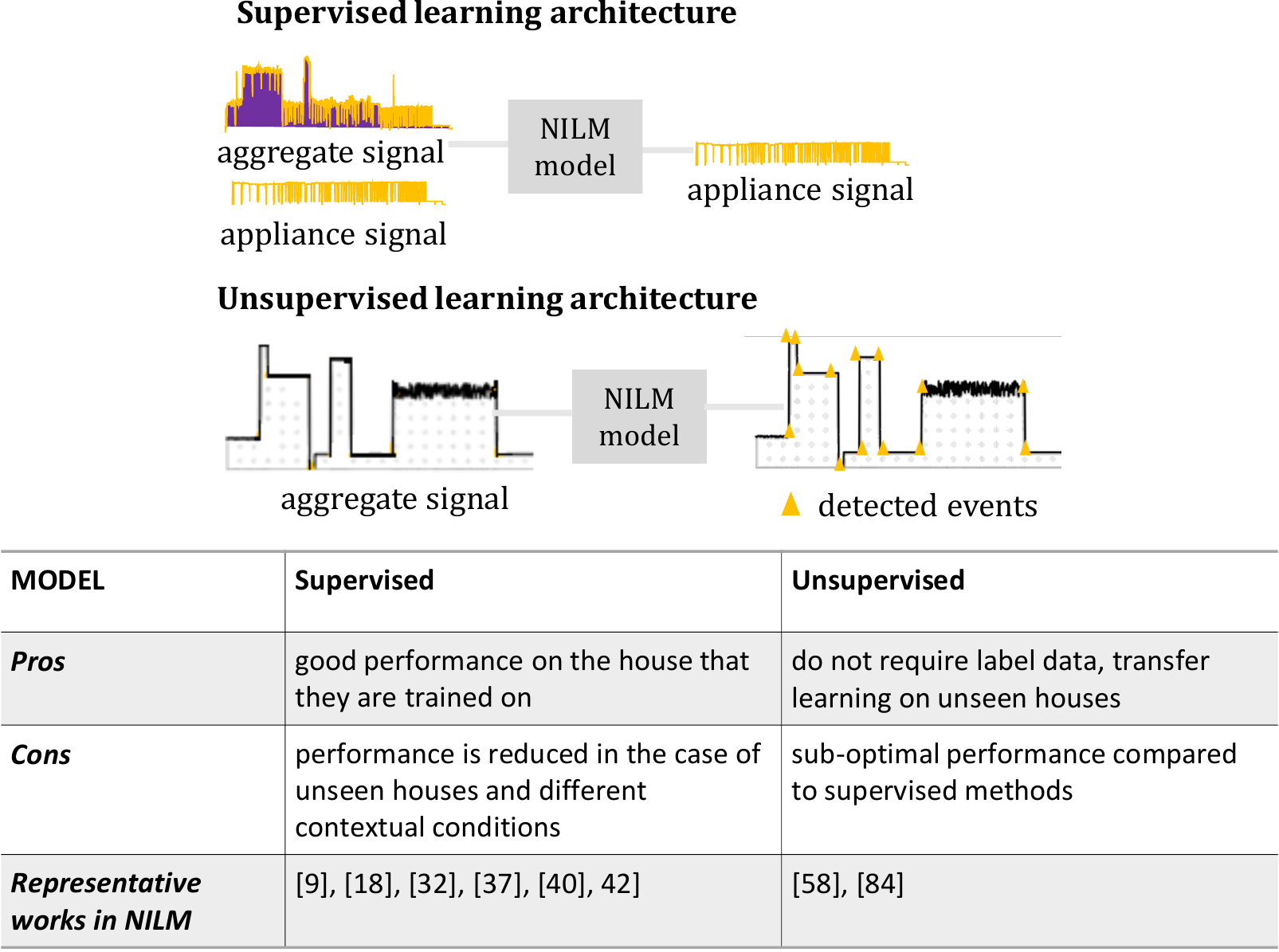}
    \caption{\textbf{Supervised versus unsupervised NILM algorithms. }}
    \label{fig:superun}
\end{figure}

Later, with the release of large amounts of labeled datasets, supervised machine learning models became trend. Nowadays, there is a variety of labeled datasets, so supervised learning is a common way of solving NILM problem. Supervised machine learning methods work with very good performance on the house that
they are trained on but are not always transferable to unseen houses \cite{khazaei2020evaluation} and different contextual conditions. On the contrary, unsupervised NILM models, usually have sub-optimal performance compared to supervised methods but they are robust to a wide range of datasets where no training information is available. This is the reason why, even if supervised deep learning methods achieve remarkable performance, the recent works propose semi-supervised \cite{barasim2015towards}, \cite{iwayemi2017saraa} or even unsupervised models \cite{jia2015fully}, in an attempt to balance between the accuracy and robustness. While supervised NILM methods are
expected to perform best on the house they were trained on, this is
not necessarily the case with transfer learning on unseen houses;
unsupervised NILM may be a better option. The advantages and disadvantages between these two approaches are summarized in Fig. \ref{fig:superun}.

\subsubsection{Convolutional or Recurrent layers}

The recent increase in availability of load data,
for model training has ignited data-driven approaches, such as deep neural networks using both convolutional neural network (CNN) and recurrent neural network (RNN) architectures \cite{Kelly2015}, \cite{Kaselimi2019a}, \cite{Murray2019}. The nature of data in load disaggregation is a uni-dimensional time series that keeps track of power consumption of each appliance in time. The NILM problem requires algorithms with the ability to process temporal information or data. 

RNNs with their recurrent connections are able to refer to previous states and therefore are suitable models for processing sequences of input data. However, RNNs lack the ability to learn long-range temporal dependencies due to the vanishing gradient problem, as the loss function decays exponentially with time \cite{Hochreiter1997Long}.
LSTMs models rely on memory cells, controlled by forget, input and output gates to achieve long-term memorization \cite{Hochreiter1997Long}. Despite their effectiveness on capturing temporal dependencies, their sophisticated gating mechanism may lead to an undesirable increase in model complexity. At the same time, computational efficiency is a crucial issue for recurrence-based models and considerable research efforts have been devoted to the development of alternative architectures, such as GRU networks. These have been widely proposed in NILM \cite{Murray2019}.

\begin{figure}
    \centering
    \includegraphics[width=8.6cm]{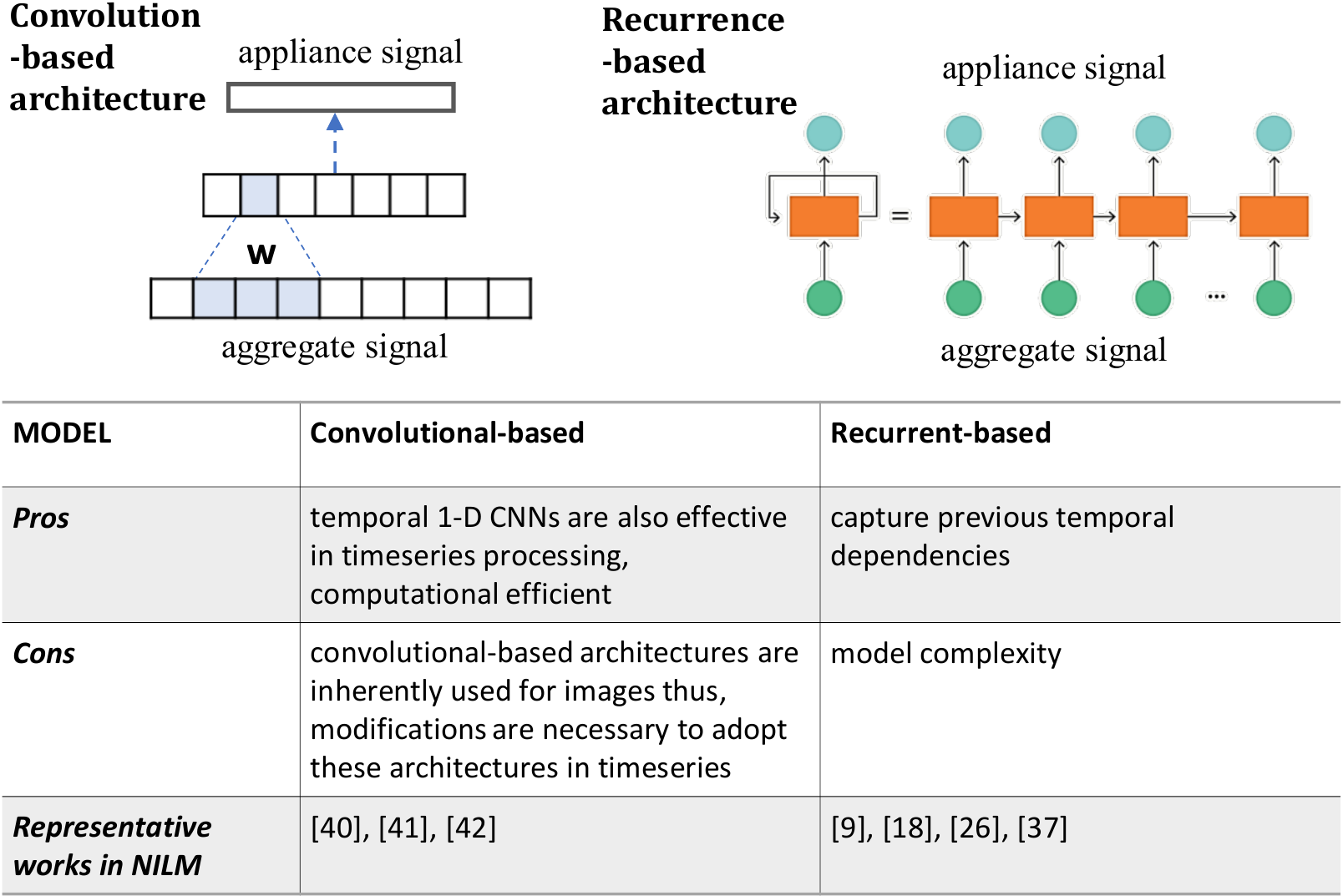}
    \caption{\textbf{Convolutional or recurrent layers for deep learning models in NILM. }}
    \label{fig:convrec}
\end{figure}

Causal or temporal \textit{1}D CNNs are also effective  in timeseries processing (see Fig. \ref{fig:convrec}). There are various works that take advantage of the emerging advancements of the traditional CNNs and their proposed modifications, to be applied in timeseries problems \cite{Oord2016WAVENET}. Thus, various works have proposed causal or temporal \textit{1}D CNN to address NILM-related challenges \cite{Harrell2019}. These networks combine causal, dilated convolutions with additional modern neural network improvements, such as residual connections and weight normalization, to reduce the required computational power without performance degradation. 

Alternative approaches suggest hybrid CNN-RNN architectures, that benefit from the advantages of both convolutional and recurrent layers. Representative examples of how these hybrid structures can be applied to NILM are \cite{Ccavdar2019new}, \cite{Kaselimi2019b}.

\subsubsection{Causal or Non causal models}

\begin{figure}
    \centering
    \includegraphics[width=8.6cm]{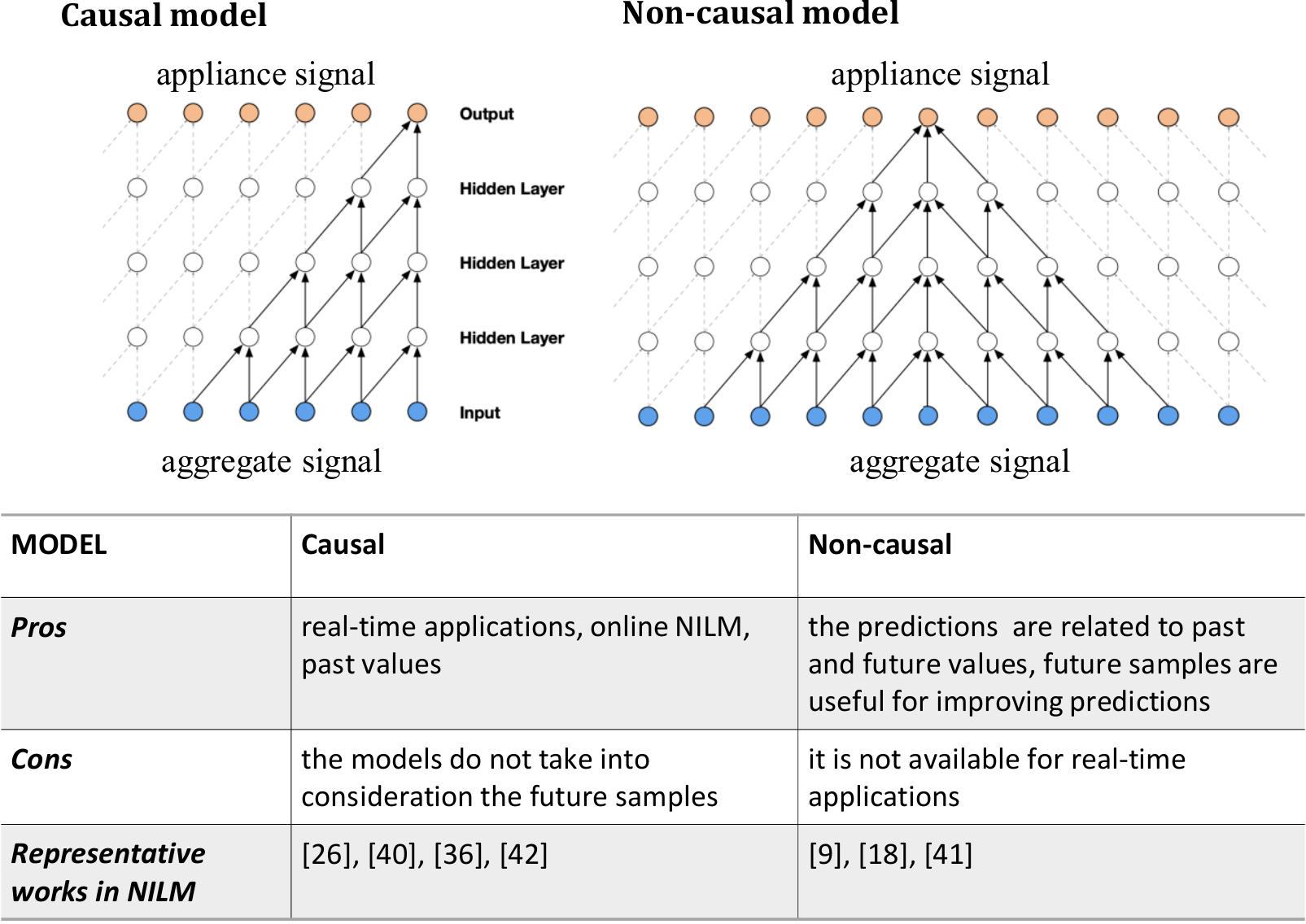}
    \caption{\textbf{The advantages and disadvantages of the causal and non-causal NILM models. }}
    \label{fig:causalnon}
\end{figure}

As indicated in Fig. \ref{fig:causalnon}, there are two different approaches applied for solving NILM based on causal and non-causal techniques. In the work of \cite{Harrell2019}, the importance of causality in NILM is highlighted. Causal convolutional neural networks use samples from previous times-steps to calculate the current output. Thus, unlike standard convolution, causal standard convolution uses the previous time step sample to predict the current result. In addition, causal dilated convolution is introduced to increase the respective field. In the causal dilated convolution the filter is applied over an area larger than its length by skipping input values with a certain step. As stated in \cite{Harrell2019}, maintaining causality is important in NILM as it allows for disaggregated data to be made available to users in real-time, achieving on-line NILM.

In cases where causality is not necessary, non-causality is important, as the future samples are generally useful for improving predictions. For a non-causal prediction, bidirectional RNN was proposed in \cite{Kaselimi2019a}, in which a backward hidden layer is added to the standard LSTM architecture to utilize the future inputs, and infer appliance's behavior, based on both past and future samples. As regards CNN-based architectures, a non-causal (bidirectional) dilated convolution is proposed in \cite{jia2021sequence}. Fig. \ref{fig:causalnon} shows that the bidirectional structures eliminate causality to access equal number of samples in the past as in the future, and makes the prediction at the center of receptive field, which results in a larger receptive field and higher performance.

\subsubsection{Sequence-to-point or Sequence-to-sequence techniques}

\begin{figure}
    \centering
    \includegraphics[width=8.6cm]{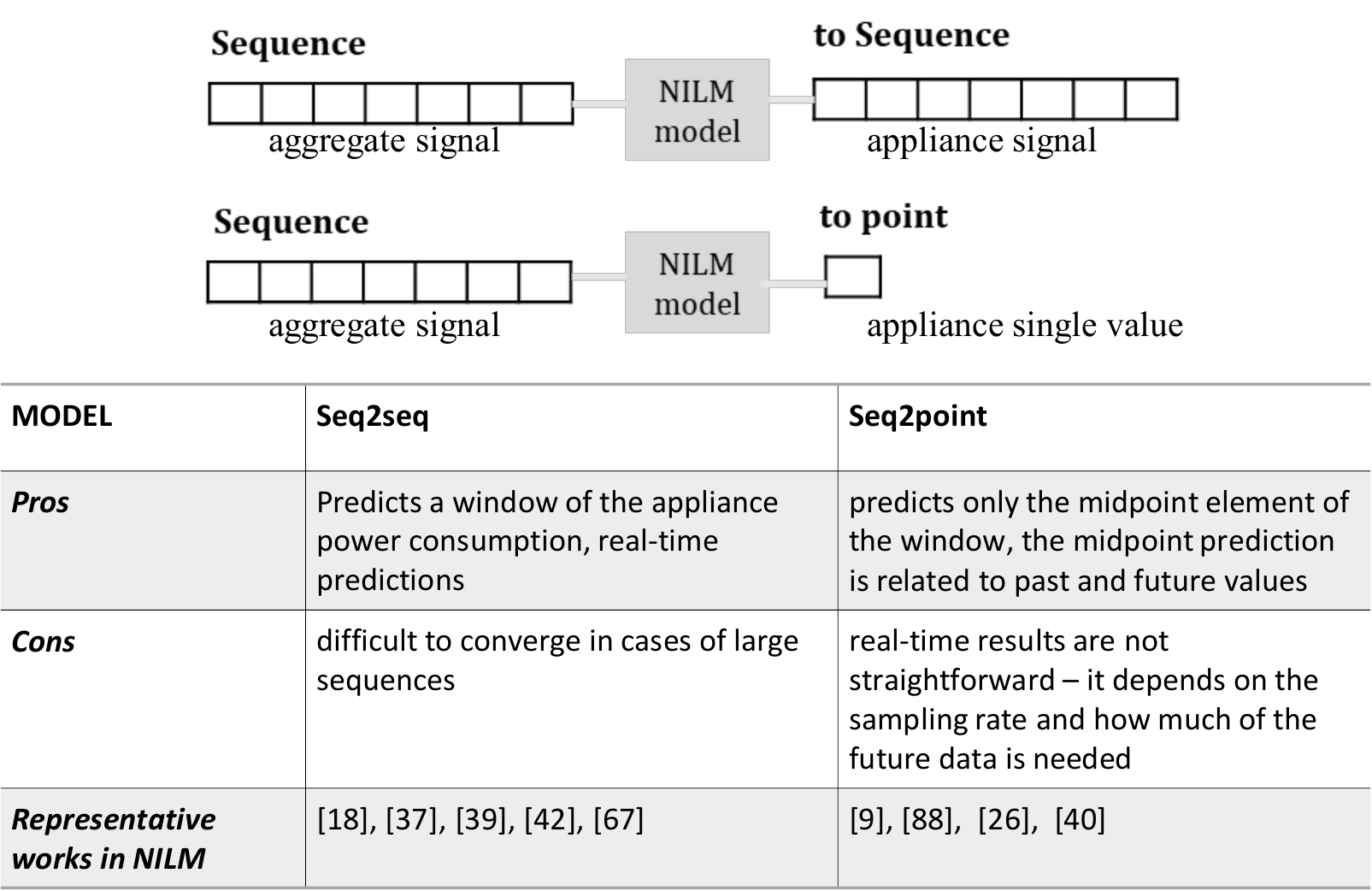}
    \caption{\textbf{Sequence-to-sequence versus sequence-to-point approaches for NILM.} }
    \label{fig:seqtoseq}
\end{figure}

The Sequence-to-Sequence (seq2seq) and Sequence-to-Point (seq2point) methods achieve remarkable accuracy results for load disaggregation tasks (see Fig. \ref{fig:seqtoseq}). Internally, they rely on neural networks, trained to identify the power consumption of a single appliance given a sequence of aggregate power data. In both methods, a window of (aggregate) input data is provided to a neural network, which has been trained to represent the relationship between the aggregate signal and the signal of the appliance under consideration. Thus, a sliding window is moved across the aggregate power signal and used to emit the disaggregated device-level power, either for sequence of the same size as the input (seq2seq) or only its mid-point element (seq2point). It is noted that due to the individual consumption characteristics of most electrical devices, a separate neural network must be trained for each device. As such, it is not strictly necessary to find a sliding window size that fits all appliances equally well. However, window size is an important parameter to be estimated. There are various works claiming that the window size is directly related to the appliance type \cite{reinhardt2020ontheimpact} and the appliance in-operation duration \cite{Kelly2015}, \cite{Kaselimi2019a}.

In cases where the length of input (aggregate) and output (appliance) sequence increases, applying seq2seq learning would make the training process difficult to converge. Seq2point learning are introduced to overcome this problem \cite{Zhang2016}. Instead of training a
network to predict a sequence of appliance power consumption values, seq2point only predicts the midpoint element of that sequence-window. This approach could make use of all nearby regions of the input sequence, past and future, making the prediction problem easier and yielding more accurate output. However, seq2point is somewhat extreme because every forward process of the model just yields as output a single value, thus introduces too much computation during the inference period. Besides, the implementation of seq2point network shows lack in accuracy \cite{Zhang2016}. 

The work of \cite{Pan2020} proposes a new perspective in the seq2seq or seq2point dilemma; a trade-off between these two approaches, i.e., the amount of computation and the difficulty level of training the neural network, by introducing a novel sequence-to-subsequence (seq2subseq) learning method. 

\subsubsection{Uni- or N-dimensional problem}

Usually, the data utilized in energy disaggregation is a uni-dimensional time series that monitors the total power
consumption at each time, along with the respective information of the load consumption of each appliance (in the case of a supervised learning approach) \cite{ayub2020multi}. Considering NILM as a time series problem, load disaggregation techniques based on sequence–to–sequence mapping are performed. Thus, given the one-dimensional input (aggregate) signal, the model learns to reconstruct the time series of a particular household appliance. Prominent examples are autoencoders as in \cite{Kelly2015}, or sequence–to–sequence algorithms with recurrent layers and their variants (LSTM, biLSTM, GRU layers) \cite{Kaselimi2020} and, more recently, temporal 1D-CNN networks \cite{Harrell2019}.

Shifting the NILM problem from uni-dimensional discrete space,  to the \textit{2}D space, is an alternative approach that has received some attention \cite{bousbiat2020exploring}. The authors of \cite{de2019detection} represent the plots of the current-voltage trajectory as binary images that are fed to a CNN-based classifier in order to identify the appliances. In the paper of \cite{schirmer2021double}, the high-frequency aggregated current and voltage signals are transformed into two-dimensional unit cells as calculated by double Fourier integral analysis and, used as input to a convolutional neural network for regression. 

In case where, the input is neither power values (\textit{1}D) nor current-voltage values (\textit{2}D), but a set of different variables such as, reactive power, apparent power, current values, etc. \cite{Kaselimi2019b}, for different appliance and houses, at different time steps then, a solution in the \textit{3}D (or \textit{N}D) space can be applied. Novel deep and tensor learning (tensor decomposition) techniques \cite{makantasis2021rank} can also be useful to decompose the total consumption into individual appliances consumption values. Batra et al \cite{batra2018transferring} propose a transferable tensor factorization approach, in which the tensor has cells that contain energy readings of the $M$ houses (\textit{1}$^{st}$ dimension) for $N$ appliances (\textit{2}$^{nd}$ dimension) and for $T$ time-steps (\textit{3}$^{rd}$ dimension).  

\subsection{Trends in machine learning approaches for solving NILM}

\begin{figure*}
    \centering
    \includegraphics[width=11.8cm]{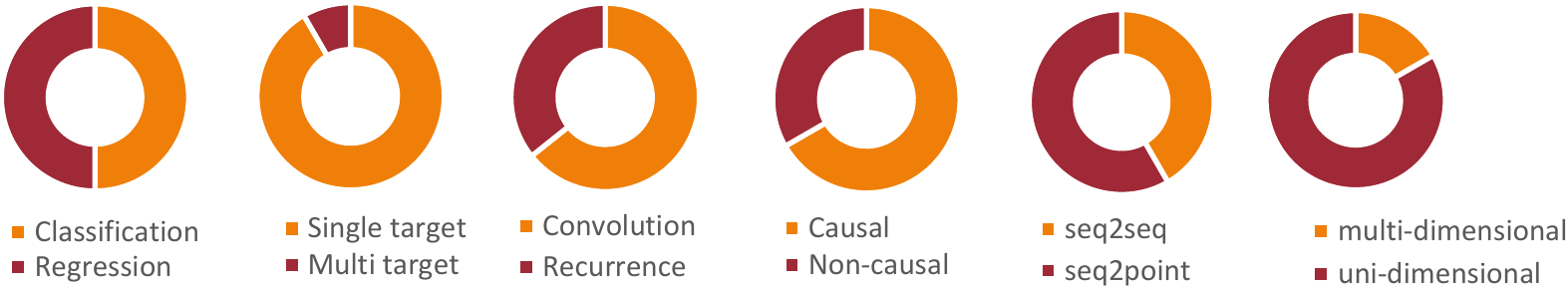}
    \caption{\textbf{Pie charts with the final choices between the different NILM research dilemas, based on the research works  referred in table 1. }}
    \label{fig:piechart}
\end{figure*}

Table 1 summarizes the most recent research works in NILM, that are dealing with the above mentioned research dilemmas and highlights the decision/proposals of each research work in respect to the research dilemmas. As regards the supervised or unsupervised research dilemma, we have already mentioned in Section \ref{subsec:opt_feat_extr}, that even though supervised learning algorithms are widely adopted in NILM, however, semi-supervised, self-supervised and unsupervised methods have recently attracted the interest of the scientific community. In this table we include only works related to supervised learning techniques. As indicated in Fig. \ref{fig:piechart} the strongest debates are taking place between the classification or regression model and the sequence-to-point or sequence-to-sequence dilemmas. As regards the single or multi target model dilemma, even though until now, the most common deep learning models in NILM are dealing with the single target approach, however, recently it has been observed an increasing interest in multi-target model approaches \cite{Buddhahai2021anon}.

\begin{table*}
\caption{\textbf{List of the representative works that target all research dilemmas and and the choices the researchers finally made. Only supervised learning techniques address all dilemmas and thus are listed in this table.}}
\label{table}
\resizebox{\textwidth}{!}{\begin{tabular}{|c|c|p{0.30\linewidth}|p{0.12\linewidth}|p{0.11\linewidth}|p{0.11\linewidth}|p{0.11\linewidth}|p{0.11\linewidth}|p{0.11\linewidth}|p{0.11\linewidth}|} 
\hline
ID&Author&Title&Classification (C)  or        Regression (R) model&Multi (M)   or         Single (S) target model&Convolutional (C) or                 Recurrent (R) based architecture&Causal (C)       or               Non-causal (N) model&seq2point or seq2seq&Uni (u) or Multi (m) dimensional \\
\hline
\hline
1& J.  Kelly  et al., 2015&Neural  nilm:  Deep  neural  networks applied  to  energy  disaggregation&Classification&Single&Convolution/ Reccurence&Non-causal&seq2seq&Uni \\
2&J., Kim et al. 2017&Nonintrusive load monitoring based on advanced deep learning and novel  signature&Classification&Single&Recurrence&Causal&seq2point&Uni\\
3&C. Zhang e. al., 2018&Sequence-to-point  learning  with  neural  networks  for  non-intrusive  load  monitoring&Regression&Single&Convolution&Causal&seq2point&Uni\\
4&K. Chen et al., 2018&Convolutional sequence  to  sequence  non-intrusive  load  monitoring&Regression&Single&Convolution&Causal&seq2seq&Uni\\
5&M.  Kaselimi et al., 2019 &Bayesian-optimized  bidirectional  lstm  regression  modelfor  non-intrusive  load  monitoring&Regression&Single&Recurrence&Non-causal&seq2seq&Uni\\
6&D.  Murray et al., 2019&Transferability  of  neural  network  approaches  for  low-rate  energy  dis-aggregation&Classification&Single&Convolution/  Reccurence&Causal&seq2point&Uni\\
7& M.  Kaselimi et al., 2019&Multi-channel  recurrent  convolutional  neural  networksfor energy disaggregation&Regression&Single&Convolution&Non-causal&seq2seq&Multi\\
8& A.  Harell et al., 2019 &Wavenilm:  A  causal  neuralnetwork  for  power  disaggregation  from  the  complex  power  signal&Classification&Single&Convolution&Causal&seq2point&Multi\\
9&M. Kaselimi et al., 2020 &Context  aware  energy  disaggregation  using  adaptivebidirectional lstm models&Regression&Single&Recurrence&Non-causal&seq2seq&Uni\\
10&A., Faustine et al., 2020 &UNet-NILM: A Deep Neural Network for Multi-tasks Appliances State Detection and Power Estimation in NILM&Classification/ Regression&Multi&Convolution&Causal&seq2point&Uni\\
11&L. d. S. Nolasco et al., 2021&DeepDFML-NILM: A New CNN-based Architecture for Detection, Feature Extraction and Multi-Label Classification in NILM Signals&Classification&Single&Convolution&Causal&seq2point&Uni\\
12&W. Yang et al., 2021&Sequence-to-Point Learning Based on Temporal Convolutional Networks for Nonintrusive Load Monitoring&Regression&Single&Convolution&Causal&seq2point&Uni\\
\hline
\end{tabular}}
\label{tab1}
\end{table*}

\section{Trustworthiness in NILM Algorithms: Can We Trust AI in NILM Problems? }
\label{sec:trust}
One major aspect of the application of AI algorithms in NILM is how reliable are their outputs, or, in other words, how one can trust the AI outputs so that one can take reliable decisions. This opens a new research field in machine learning society, called \textit{trustworthy AI}. NILM ML-based models should be reliable in order to gain consumers' trust, otherwise NILM-based technologies won't enable the consumers' to change their behavior. In this section, we discuss the key papers towards {\it trusted NILM solutions}. A trustworthiness solution of an AI algorithm refers to six main aspects \cite{Liu2021trustworthy}; reliability, scalability, robustness, explainability, fairness and privacy of the NILM algorithms.  

\subsection{Reliability} 
In terms of NILM, reliability implies that the model could accurately distinguish the similar appliances, and avoid the probability of mis-classification and  misinterpretation of results. NILM models reliability is even more crucial in cases of faulty appliances detection or for real-time applications. In order to achieve this, there are a few works in literature that propose optimization techniques for NILM deep learning models that fine tune the algorithms to accurately detect the appliances in operation with the minimum error \cite{Kaselimi2019a}, \cite{Faustine2021adaptive}. The adoption of online learning techniques is necessary for the NILM algorithms to dynamically adapt to new patterns in the appliances data and in contextual changes (e.g., related to environment or seasonal changes). Changes maybe referring to: \textit{i)} appliance faulty operation, \textit{ii)} different types of appliances models (model testing in unseen houses), \textit{iii)} changing of appliance operation due to changing environmental conditions and seasonality, \textit{iv)} entrance of a new domestic appliance in the total load. The work of \cite{Kaselimi2020} proposes a context-aware model that is adapted in the various conditions, resulting into an improved performance compared to traditional deep learning NILM models that are trained only once.

Under a NILM framework, it is important to deploy flexible NILM algorithms, adaptable to new appliances or appliance replacements, as in \cite{jones2020stop} and \cite{salem2019semi}. Here, it is worth-mentioned the work of \cite {salem2019semi} that proposes a semi-supervised approach for online learning for NILM using conditional Hidden Markov Models (HMM). In order to accurately detect all these changes, continual/active learning methods are essential, given that stationary pre-trained models cannot effectively deal with non-stationary appliances power data distributions. As regards the transferability of NILM algorithms, i.e., the ability of the algorithm to  disaggregate appliance loads which have previously not been seen (or trained) by the NILM
solution\cite{Murray2019},  has been widely studied in various research works such as in \cite{Murray2019}, \cite{DIncecco2019}, \cite{klemenjak2019onmetrics}. 

\subsection{Scalability} 

Scalable AI for NILM is defined as the ability of (i) data, (ii) NILM algorithms, and (iii) infrastructure to operate at the size, speed, and complexity required to solve the NILM problem. The challenge of making NILM models scalable is crucial mostly because the existing deep learning solutions resulting in models with millions of parameters and a high computational burden. On the contrary, the utilities should perform a large-scale deployment to support thousands of consumers to benefit as much as possible from energy disaggregation services. Here, it is important to highlight that multi-target models (see Section \ref{subsection:multitar}) usually suffer from scalability problem as the number of devices to observe rises, and the inference step is computationally heavy. The current state-of-the-art NILM algorithms propose efficient techniques and models that do not require vast amounts of trainable parameters \cite{Faustine2021adaptive}. Secondly, the proposed system should be delay-free; once the appliance has been turned-on, the system is able to calculate its power in near real-time. A scalable real-time event-based energy disaggregation methodology using convolutional neural network is proposed by Athanasiadis et al. \cite{athanasiadis2021scalable}, whereas Krystalakos et al. \cite{krystalakos2018sliding} propose a real-time energy disaggregation method based recurrent network architectures.  

\subsection{Robustness} 

The power signal exhibits severe non-linearity, since the temporal periodicity of the individual appliance activation depends on contextual characteristics \cite{batra2015neighbourhood}, i.e., geographic and socioeconomic parameters or even residents habits. This leads to diverse energy consumption patterns in households. Therefore, it is challenging to implement models with good generalization ability that achieve high performance when tested on unseen houses. The importance of the number of houses can be explained in two ways. First, machine learning
approaches for NILM can have an overfitting problem when the number of houses is not large enough.
Data acquired from many houses can be crucial for a better generalization of NILM algorithms. As the
number of houses increases, the number of combinations of appliances covered by the algorithm
also increases, which makes NILM algorithms applicable to new houses. Secondly, the diversity of
models for the same appliance type cannot be addressed by the limited datasets that are open access available \cite{shin2019data}. Kaselimi et al. \cite{Kaselimi2021} propose a GAN-based framework for NILM that is robust even in presence of noisy data input, achieving better results compared to other traditional deep learning models. Welikala et al. \cite{Welikala2016} propose a NILM method, that is robust even in the presence of unlearned or unknown appliances.
In \cite{rafiq2021generalizability}, a data augmentation technique is proposed in order to improve the generalization ability on new unseen data. The technique  combines on and off-duration of the target appliance from various datasets, to form a synthetic aggregate and sub-meter profiles. 

\subsection{Explainability} 

Understanding the outputs of the networks contributes in improving the NILM model structure, highlights the relevant features and aspects of the data used for making the decision, provides a more clear picture of the accuracy of the models (since a single accuracy number is often insufficient), and also inherently provides a level of trust in the value of the provided consumption feedback to the NILM end-user. Murray et al. \cite{murray2020explainable}, \cite{Murray2021transparent} investigate how eXplainable AI (XAI-based) approaches can be used to explain the inner workings for NILM deep learning models, and examine why the network performs or does not perform well in certain cases. Explainable AI is utilized to analyse input data and address biases, especially when the NILM algorithms are tested in unseen houses in order to improve the performance of the models \cite{Murray2021transparent}.

\subsection {Fairness} 

In literature, fairness in AI is defined as the absence of prejudice or preference for an individual or a group based on their characteristic attributes \cite{mehrabi2021survey}. Bias exist in many shapes and forms, and is introduced at any stage of the model development pipeline. As regards data interpretation, the datasets usually suffer from historical bias, representation bias, measurement bias, temporal bias or even omitted variable bias. As regards, the machine learning model deployment, the algorithmic and evaluation biases are usually met in NILM techniques. Finally, human-in-the-loop approaches that consider human as reviewer of the model's predictions can also introduce their own biases, when they decide whether to accept or disregard a model’s prediction. Examples of these kind of biases are social or behavioral biases.

\subsection {Safety and Privacy} 

Deep learning based NILM models largely rely on sufficient and diverse training data gathered in centralized platforms. Even though there are plenty of meter data in different households, it is almost impossible to transmit or integrate these local user data into a centralized storage, due to limits arising from the legislation in consumer privacy and data security. In addition, even though smart meter devices collect data with high resolution, however, the storage is usually performed every 15 minutes for the integral of consumption for privacy-preservation reasons. As a result, disaggregating timeseries with a timestep of 15 minutes is  much more demanding in these cases. Data security and user privacy issues have become a an issue of major importance. Thus, privacy and security of sensitive consumers' data should be enhanced at all levels of the data processing workflow. Safety and privacy issues may imply fewer data available, therefore additional challenges for training NILM models with good performance are appeared.
However, recently federated schemes have been merged as the state-of-the-art techniques in order to achieve personalized energy disaggregation with the state-of-the-art accuracy, while ensuring privacy preserving for the consumer \cite{Zhang2021fednilm}. These schemes allow for federated model training, without requiring data transfer and safeguarding that the data does not have to leave the local source premises. 

\section{Datasets, Performance Evaluation/Validation Strategy and Open NILM Tools}
\label{sec:evaluation}

\begin{table*}
\caption{\textbf{An overview of NILM datasets. The table summarizes the year of the release per dataset, the number of houses includded as well as the duration of the dataset and the measured variables. Also the last common refers worth mentioning notes for each dataset.}}
\begin{center}
\resizebox{\textwidth}{!}{\begin{tabular}{|m{0.07\linewidth}|m{0.025\linewidth}|m{0.07\linewidth}|m{0.03\linewidth}|m{0.08\linewidth}|m{0.12\linewidth}|m{0.08\linewidth}|m{0.08\linewidth}|m{0.23\linewidth}|}
\hline
Dataset Name&Year&Country&House Num.&Duration&Variables&Aggregate sampling rate&Appliance sampling rate&Comments\\
\hline\hline
\textbf{REDD \cite{REDD}}&2011&USA&6&a few months&current, voltage&1Hz, 15kHz&1/3Hz&first released and most-used\\
\hline
\textbf{BLUED \cite{BLUED}}&2011&USA&1&8 days&current, voltage&12 kHz&-&allows for analysis in both the time and the frequency domains\\
\hline
\textbf{HES \cite{HES}}&2012&UK&251&1 year&active power&2-10min&2-10min&number of houses\\
\hline
\textbf{AMPds \cite{Makonin2013}}&2013&Canada&1&1 year&current,voltage, pf, real, reactive and apparent power&1 min&1 min&multiple variables\\
\hline
\textbf{BERDS \cite{BERDS}}&2013&USA&1&1 year&active, reactive and apparent power&20s&20s&public building on the University\\
\hline
\textbf{iAWE \cite{iAWE}}&2013&India&1&73 days&current, voltage, active, reactive and apparent power&1s&1s&contains electricity, gas, and water consumption data\\
\hline
\textbf{DRED \cite{DRED}}&2014&Netherlands&1&6 months&active power&1Hz&1Hz& indoor and outdoor temperature, wind speed, humidity, precipitation, occupancy information \\
\hline
\textbf{ECO \cite{ECO}}&2014&Switzerland&6&8 months&active reactive power&1Hz&1Hz&occupancy information of the monitored household\\
\hline
\textbf{GREEND \cite{GREEND}}&2014&Italy/Austria&9&1 year&active power&1s&1s&cross-countries dataset \\
\hline
\textbf{PLAID \cite{PLAID}}&2014&USA&60&summer of 2013 and winter of 2014&current, voltage &-&30 kHz&3 versions, PLAID 1(2014), PLAID 2 (2017) and PLAID 3 (2018)that includes also aggregate measurements\\
\hline
\textbf{REFIT \cite{Murray2017}}&2015&UK&20&2 years&active power&8 s&8 s&corrupted with noise version of the dataset\\
\hline
\textbf{UK-DALE \cite{kelly2015theukdale}}&2015&UK&5&1 to 2.5 years&current, voltage&6s, 16kHz&6s&long duration\\
\hline
\textbf{COOLL \cite{COOLL}}&2016&France&1&-&current, voltage&-&100kHz&High-Sampled Electrical Signals for Appliance Identification   \\                                                         \hline             
\textbf{BLOND \cite{BLOND}}&2018&Germany&1&213&current, voltage&50 kHz&6.4 kHz&building-level office environment dataset\\
\hline
\textbf{EMBED \cite{EMBED}}&2019&USA&3&14-21 days&active, reactive power&12 kHz& 12 kHz&aggregate power files, fully labeled appliance event timestamps, and plug load consumption for a variety of monitored appliances\\
\hline
\textbf{SynD \cite{SynD}}&2019&Austria&1&180 days&active power&5Hz&5Hz&synthetic energy dataset\\
\hline
\textbf{DEDDIAG \cite{DEDDIAG}}&2021&Germany&15&<3.5 years&active power&1Hz&1 Hz&long duration\\
\hline
\textbf{IDEAL \cite{IDEAL}}&2021&UK&255&<2 year&active power&1s&1s&electricity and gas sensor data along with a diverse range of relevant contextual data from additional sensors and surveys\\
\hline
\end{tabular}}
\end{center}
\end{table*}

\begin{table*}
\caption{\textbf{NILM research from a practical perspective. Datasets utilized, performance achieved per appliance and adaptability of the NILM algorithms. In cases where unseen houses option is checked, the performance in the table is referred to unseen houses of the same dataset that are used only for training.}}
\begin{tabular}{|c|c||ccccc||p{0.1\linewidth}|p{0.04\linewidth}|}
\hline
 & & \multicolumn{5}{c}{Top-5 common appliances MAE [W]}&&\\
\hline
A/A&Dataset&dishwasher&Washing Mach.&Fridge&Microwave&kettle& Overall MAE [W] per dataset&Unseen house\\
\hline
\cite{Kelly2015}     &UK-DALE          &24.0& 11.0 &18.0&6.0&6.0 &22.0&\checkmark\\
\cite{Zhang2016}     &REDD/UK-DALE     &20.0/27.7 &18.4/12.6&28.1/20.9&28.2/8.7&-/7.4&15.5 /23.6&\checkmark\\
\cite{Chen2018}      &REDD       &12.8  &\cellcolor[HTML]{E1E4E7}&32.0&\cellcolor[HTML]{E1E4E7}&\cellcolor[HTML]{E1E4E7}  &-&\checkmark\\
\cite{Kaselimi2019a} & AMPds           & \textbf{6.4}& 9.2& \cellcolor[HTML]{E1E4E7}& \cellcolor[HTML]{E1E4E7}& \cellcolor[HTML]{E1E4E7} & -& \\
\cite{Murray2019}    &REDD/REFIT       &119.4/82.74&-/71.9&10.1/\textbf{8.6}&68.0/35.5   &\cellcolor[HTML]{E1E4E7}&-&\checkmark\\
\cite{Kaselimi2019b} &AMPds            &14.3&\textbf{4.8}&\cellcolor[HTML]{E1E4E7}&\cellcolor[HTML]{E1E4E7}&\cellcolor[HTML]{E1E4E7}&-&\\
\cite{shin2019subtask}&REDD/UK-DALE     &15.9/13.5 &20.6/11.0&22.9/15.3&15.9/8.6  &\cellcolor[HTML]{E1E4E7} & 18.8 /10.9&\checkmark\\
\cite{Kaselimi2020}  &REDD/REFIT/AMPds &7.1/31.3/-&-/21.8/9.2   &\cellcolor[HTML]{E1E4E7}& 6.9 /-/-&\cellcolor[HTML]{E1E4E7}        &-&\checkmark\\
\cite{faustine2020unet}  & UK-DALE         &6.8&11.5    &15.2&6.5&16.0  &11.2&\checkmark\\
\cite{DIncecco2019}       &REDD/UK-DALE/REFIT          &20.0/27.7/12.2   &18.4/12.6/16.9 &28.1/20.9/20.0&28.2/8.7/12.7&\cellcolor[HTML]{E1E4E7}  &23.7/15.5/13.7&\checkmark\\
\cite{Yang2021}      &UK-DALE          &23.3      &\cellcolor[HTML]{E1E4E7}    &16.4&12.6 &4.1&-&\checkmark\\
\cite{Pan2020}       &UK-DALE          &13.5      &7.1   &11.9& \textbf{3.1}&\textbf{3.6}   &7.8&\checkmark\\
\cite{lin2021deep}       &REDD/UK-DALE/REFIT          &23.8/28.4/15.4&19.9/15.9/17.9&31.3/22.3/23.2&29.9/9.7/12.2&-/7.7/6.9&-&\checkmark\\
\cite{liu2021unsupervised}       &UK-DALE/REFIT          &51.3/28.2&25.9/44.0&31.1/63.1&64.5/20.7&13.9/16.7&-&\checkmark\\
\cite{yue2020bert4nilm}       &REDD/UK-DALE      &20.5/16.2     &34.9/6.9    &32.4/25.5&17.6/6.96&-/6.8&26.4/12.4&\checkmark\\

\hline
\end{tabular}
\end{table*}

\subsection{Datasets}
The existing surveys for NILM, highlight the importance of selecting the right dataset. Huber et al. \cite{huber2021review} summarize the main characteristics of the open access data sets.  In addition, in \cite{klemenjak2020towards} a comparison between the different open access datasets is performed. Table 2 describes the most common datasets used for NILM along with their characteristics. Here, we highlight that the selection of a dataset is directly related to the NILM method you would follow, since a dataset sets its own constraints, preventing the application of some machine learning models. For example, as indicated in Table 2, AMPds \cite{Makonin2013} dataset, has a sampling rate of 1 min. This means that this dataset cannot be used in high-frequency applications, but only for the low-frequency ones. On the contrary, BLOND \cite{BLOND} and EMBED \cite{EMBED} datasets are relevant for high-frequency models. Another attribute is the overall data time duration in each dataset. The largest ones have duration for few years (i.e., REFIT \cite{Murray2017}, DEDDIAG \cite{DEDDIAG} and IDEAL \cite{IDEAL}), whereas the smaller ones only for few days, as it is the case in REDD dataset \cite{REDD}. Also, most of the datasets listed in Table 3, present limited number of houses. On the contrary, most of the current techniques for NILM require for their training a significant amount of labeled appliances power data. However, such a collection which is a major bottleneck in developing robust and generalized NILM solutions. Exceptions are the HES \cite{HES}, PLAID \cite{PLAID} and IDEAL \cite{IDEAL} datasets. Recently, with the progress of machine learning, Generative Adversarial Networks (GANs) are employed to synthesize appliance power signatures \cite{harell2021tracegan}, \cite{renaux2020dataset}. SynD \cite{SynD} is an example of synthetic dataset.

\subsection{NILM metrics and evaluation}
Until now, there is no common understanding or accepted format on how to report the accuracy results in NILM \cite{klemenjak2020towards}. However, there are a few research works that present an overview of performance evaluation metrics in NILM and the worth-mentioning ones are the references of \cite{makonin2015nonintrusive}, \cite{pereira2019nilmpeds},  \cite{pereira2017comparison}, \cite{klemenjak2020towards}. Under a NILM framework, the metrics need to be reported in overall disaggregation scores (household-level) and appliance specific scores (appliance-level). The household-level metrics show the overall model accuracy and capability of disaggregating the total power signal into its component signals. This type of evaluation could be beneficial in application scenarios where a big set of houses is incorporated into a single NILM model. In this scenario, one should test model's accuracy at-a-house-level and investigate model's transferability in different houses, through the evaluation of model's performance in each house separately. Also, it is important to report the performance per appliance (appliance-level metrics) in order to identify strengths and weaknesses of the different NILM algorithms. With this more detailed accuracy information, one could imagine a system that would select different algorithms depending on the context (including specific history) of the disaggregation task.  
As a consequence of the variety of existing load disaggregation techniques, performance evaluation has to objectively assess classification as well as regression performance in order to enable comparability. Thus, metrics utilised in NILM can be divided into: (i) classification metrics used to evaluate model's performance on an appliance event detection (e.g., ON-OFF events) and, (ii) power estimation metrics applied to regression based approaches assessing model's performance on appliance power signal decomposition from the total power signal \cite{klemenjak2020towards}. 

\subsubsection{Classification -event detection- metrics} 

The performance evaluation in this approach aims to assess method effectiveness in the accurate estimation of appliance status (on/off events). Here, the most common metrics used are: the accuracy, the precision, the recall and the F1-score \cite{pereira2019nilmpeds, pereira2017comparison}. In particular, Pereira et al. \cite{pereira2017comparison} analyse experimentally the behaviour of eighteen different performance metrics applied to classification NILM algorithms.

\subsubsection{Regression -power estimation- metrics} 
The performance evaluation of this category (regression-based approaches) aims to assess the effectiveness of a method by comparing the observed appliance signal (ground-truth) and provided estimates. Common used metrics are: the mean absolute error and the root mean squared error. Pereira et.al in their work provide a detailed list of metrics used along with their description \cite{pereira2017comparison}. 

Table 3 presents a comparison between the common metrics used for the evaluation of the NILM algorithms in various research works. MAE is selected for comparison purposes because it is the most common metric in the bibliography up to now. The second column indicates the datasets where the comparisons are performed. Table displays the performance errors (MAE in Watt units) for the 5 most common appliances in disaggregation tasks. In bold, we highlight the minimum MAE error in which the proposed model achieves the best performance. Grey cells indicate that the disaggregators for a specific appliance are not available. The last column marks in tick the literature works in which the models are tested for unseen houses of the same dataset. The work of \cite{Kaselimi2019a} proposes a bayesian-optimized bidirectional LSTM network that achieves the minimum MAE error for the dishwasher appliance in AMPds dataset. As regards the washing machine appliance, the minimum MAE error was achieved in \cite{Kaselimi2019b}. Murray et al. \cite{Murray2019} propose a convolutional based network that achieves the highest performance in the fridge appliance for REFIT dataset. As regards the microwave and kettle appliances. Pan et al. propose a model based on conditional GANs with the minimum error for these appliances in UK-DALE dataset, i.e., 3.1 and 3.6 W, respectively. Therefore, it is clear that disaggregation performance highly depends on appliance type.

\subsection{Open NILM tools towards Commercialization}

NILMTK \cite{Batra2014}, and its recently released version NILMTK-Contrib \cite{kukunuri2020nilmtk}, is a common tool and the most well-known framework for bench-marking in NILM towards reproducible NILM algorithms. Its presence as an open-access toolkit, and the successful implementation of various energy disaggregation algorithms, has unfolded the means for comparisons of the different algorithms in the NILM research community. Also, it enables researchers to observe and evaluate NILM approaches in multiple datasets accessible online. Except for comparing against benchmarking disaggregation algorithms, NILMTK provides dataset statistics, pre-processing tools and NILM metrics that help towards comparability of the various proposed methods. 

Except for NILMTK, which is a combined effort towards reproducible and standardization of NILM algorithms, there are several works and NILM researchers that publish the code associated with their works on code repositories to be free available. Worth mentioning research works that have been published with source code or extensive supplemental material are the following: the WaveNILM model \cite{Harrell2019}, the Neural NILM model \cite{Kelly2015} and the seq2point NILM model \cite{Zhang2016}. 



\section{Discussion and Conclusion}
\label{sec:discussion} 
NILM is an important process for various reasons. Energy disaggregation is an essential element of energy conservation, efficiency and careful energy utilization, since it elaborates on the energy usage tendencies. The trends observed in energy usage patterns from a household can be used for security purposes, given that an anomaly might implies appliance failure or illegal use of supplied electricity. The appliance usage patterns can also be used to calculate and control the amount of carbon emissions. 

Having identified and highlighted the most important research works in NILM and critically analysing the information gathered by identifying the gaps in NILM, as final step, we summarize the most important aspects in the NILM pipeline in order to implement effective NILM algorithms. In particular, we discuss issues related to i) the pre-processing and feature extraction and selection  phase, ii) the NILM model implementation phase and, iii) the model evaluation phase. 

\subsection{Discussion on feature selection and data pre-processing in NILM}
\label{subsec:best_prectices}

Aggregated (total) power consumption data stemming from smart metering measurements, are essential in order to implement a successful NILM algorithm. Assuming a supervised NILM scheme, except for the aggregated power measurements, additional information of appliances' power consumption signals is also collected. The optimal features extracted framework of Section \ref{subsec:opt_feat_extr} can be also utilized to improve the model performance. We highlight that the most common additional input variables reported in literature are: i) the time (e.g., temporal resolution), ii) spatial information (e.g., household's location), and iii) the events (e.g., appliances availability, local regulations, existence of Photo-Voltaic (PV) panels). These additional variables are ingested into the model to increase its performance. However, most works use only a limited subset of these variables \cite{sadeghianpourhamami2017comprehensive}. Selecting the most effective input variables for a machine learning NILM model is a challenging problem (see Section \ref{sec:trust}) and is closely related to model performance and computational cost, as discussed in \cite{sadeghianpourhamami2017comprehensive}. To address these issues, some works propose a feature elimination process to identify the most effective feature set \cite{sadeghianpourhamami2017comprehensive}. However, selecting a set of optimal features is constrained from the data availability and is closely related to the methods that will be used for appliances' identification \cite{bao2020feature}.

Regarding data sampling period, a coarse division of 1$s$ usually enables feature separation between macroscopic (low-frequency) and microscopic (high-frequency) components. Even though, low-frequency data sets greatly reduce the ability to distinguish among different types of appliances, compared to high- frequency data sets, the first are gaining more and more ground in NILM deep learning algorithms  \cite{huber2021review}. 

Another important step for a successful NILM algorithm is to handle the missing data due to various causes such as metering and transmission equipment failures. In addition, large and sparse outliers, occurring out of transients, surges and non-linearities in the load, are also observed \cite{Gupta2016Handling}. To handle these issues, we can apply: i) cluster-based handling (CBH), ii) interpolation, or iii) omitting specific entries. All these approaches have some disadvantages that need to be considered. At first, CBH could be time consuming and is sensitive to the parameters. A centroid-based clustering (k Nearest Neighborhood approach-kNN) could solve these issues but it cannot manage the outlier case. Yet, a density-based clustering (e.g., the OPTICS) will tackle the outlier case but it may fail  on large data sets due to numerical instabilities. Secondly, interpolation approaches (linear or not) is a common approach for filling short gaps in power timeseries data. If those assumption do not hold, then we end with miss-informative cases. Finally, omitting entire entries can be used while preparing a train and validation set, but this approach cannot solve the on-site monitoring request.

\subsection{Discussion on NILM model implementation}

NILM methods are insufficiently mature, and their performance varies on the datasets used, or the appliance that is selected for disaggregation. Table 3 shows that the model proposed by \cite{Pan2020} performs best for microwave and kettle appliances. However, in case of dishwasher and washing machine appliances, there are other approaches that achieve better performance. Usually, while an implemented algorithm may be ideal for an appliance, its performance may not be good enough for another. Thus, we cannot claim that there is a holistic approach which is the ‘best’ for any appliance type.

Regardless of the machine learning-based strategy that each researcher follows, based on the research dilemmas that are summarized in Section \ref{sec:dl4nilm}, NILM researchers should have in mind that the most important aspects in NILM arise from the practical issues encountered in this demanding domain. Machine learning and deep learning community proposes an ever-increasing number of algorithms with advanced capabilities, however, with increasing complexity. It is a considerable challenge to propose efficient and practical state-of-the-art NILM algorithms that in addition,  comply with the requirements of trustworthiness, as discussed in Section \ref{sec:trust}, forming feasible solutions towards commercialization. The newly-introduced transformer models with attention layers for NILM require a lot of computational resources and scenarios with continuous learning mechanisms arise computational complexity issues. Recently, there are proposed models based on transformers architectures that introduce techniques for  computationally efficient transformers \cite{sykiotis2022electricity}.

\subsection{Discussion on NILM model evaluation}

Despite a big variety of performance measures that are observed in the literature (see Section \ref{sec:evaluation}), it is crucial to select metrics carefully in order to avoid misinterpretations of results \cite{klemenjak2020towards}. It is important also, to keep in mind when reporting the accuracy metric, that the results needs to be normalized. Normalization of the results allow the readers to understand the relative standings from one appliance to another and from each appliance to the overall accuracy. In addition, reporting specific scores for appliance states is not necessary because different models of appliances follow a different event detection method and divide the states in a different way. Thus, based on the method that a researcher applies, different number of states at different power levels are extracted. Consequently, this metric is not comparable. Thus, it is important for the NILM researchers to present both classification and regression metrics in their studies, regardless whether NILM has been previously formulated as a regression or classification problem.

\bibliographystyle{IEEEtranN}
\bibliography{refs}

\begin{IEEEbiography}[
{\includegraphics[width=1in,height=1.25in,clip,keepaspectratio]{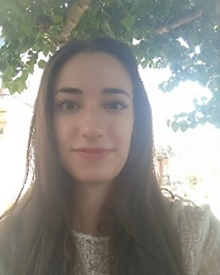}}]{Maria Kaselimi} received the Diploma, MSc and PhD degrees from National Technical University of Athens (NTUA). Her research interest focuses on machine learning, signal processing techniques, data analysis and modeling with applications in the fields of earth monitoring and environment. She has 25 papers in international journals and conferences and more that 160 citations. She has been involved in several European research projects, as researcher. 
\end{IEEEbiography}

\vfill

\begin{IEEEbiography}[
{\includegraphics[width=1in,height=1.25in,clip,keepaspectratio]{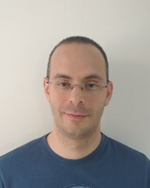}}]{Eftychios Protopapadakis}  studied production engineering and management at Technical University of Crete. His educational background includes a M.S. degree in management and business administration and a Ph.D. in decision systems, both at the same university. He has worked as engineer in European (4D-CH-World, BENEFFICE, eVACUATE, ROBO-SPECT, Terpsichore, WaterSpy) and Intereg (e-Park, Poseidon) projects since 2010. His research interests focus on machine learning applications. He has explored the applicability of semi-supervised techniques in maritime surveillance, energy applications, elder people support, industrial workflow monitoring, structural assessment of tunnel infrastructures, and cultural heritage applications. Eftychios has co-authored more than forty publications, receiving more than 2500 citations.  
\end{IEEEbiography}
\vfill

\begin{IEEEbiography}[
{\includegraphics[width=1in,height=1.25in,clip,keepaspectratio]{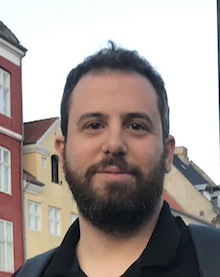}}]
{Athanasios Voulodimos} (Member, IEEE) received the Dipl.-Ing., M.Sc., and Ph.D. degrees
from the School of Electrical and Computer Engineering
of the National Technical University of
Athens (NTUA) ranking at the top of his class. He
is an Assistant Professor with the School of Electrical and Computer Engineering at NTUA. From 2018 to 2021 he was an Assistant Professor at the Department of Informatics and Computer Engineering of the University of West Attica. He has been involved
in several European research projects, as a Senior
Researcher and a Technical Manager. He was recipient
of the awards for his academic performance
and scientific achievements and has coauthored more than 120 papers in international
journals, conference proceedings and books in the research areas of
machine learning and signal processing, including their applications in earth sciences, energy and environmental engineering, receiving more than 3000 citations.
\end{IEEEbiography}
\vfill

\begin{IEEEbiography}[
{\includegraphics[width=1in,height=1.25in,clip,keepaspectratio]{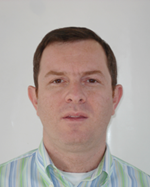}}]{Nikolaos Doulamis} (Member, IEEE) received
the Diploma and Ph.D. degree in electrical
and computer engineering from the National
Technical University of Athens (NTUA) both with
the highest honor. He is currently an Associate
Professor with the NTUA. He has received many
awards (e.g., Best Student among all Engineers,
Best Paper Awards). He was an Organizer and/or
TPC in major IEEE conferences. He has authored
more than 75 (240) journals (conference) papers in
the field of signal processing and machine learning, many of them with applications in geoinformatics and earth sciences,
and received more than 7800 citations. He has been involved in several European research projects.
\end{IEEEbiography}
\vfill

\begin{IEEEbiography}[
{\includegraphics[width=1in,height=1.25in,clip,keepaspectratio]{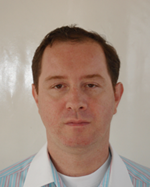}}]{Anastasios Doulamis} (Member, IEEE) received the Diploma and PhD degree in Electrical and Computer Engineering from the National Technical University of Athens (NTUA) with highest honor. Until January 2014, he was an Associate Professor at the Technical University of Crete and now is an Assistant Professor at NTUA. He has received several awards in his studies, including the Best Greek Student Engineer, Best Graduate Thesis Award. He has also served as program committee in several major conferences of IEEE and ACM. He is author of more than 350 papers in leading journals and conferences receiving more than 7541 citations. 
\end{IEEEbiography}

\EOD

\end{document}